\pdfoutput=1
\documentclass[11pt]{article}

\usepackage{acl}

\usepackage{times}
\usepackage{latexsym}

\usepackage[T1]{fontenc}
\usepackage[utf8]{inputenc}

\usepackage{microtype}

\usepackage{booktabs}
\usepackage{graphicx}
\usepackage{amssymb} % For checkmark symbol
\usepackage{multirow}

\title{R.U.Psycho? Robust Unified Psychometric Testing of Language Models}

\author{Julian Schelb\\
  University of Konstanz \\
  \texttt{julian.schelb@uni.kn} \\\And
  Orr Borin \\
  Recosys \\
  \texttt{orr.borin@recosys.com} \\\AND
  David Garcia \\
  University of Konstanz \\
  \texttt{david.garcia@uni.kn} \\\And
  Andreas Spitz \\
  University of Konstanz \\
  \texttt{andreas.spitz@uni.kn} \\
}

\begin{document}
\maketitle
\begin{abstract}
Generative language models are increasingly being subjected to psychometric questionnaires intended for human testing, in efforts to establish their traits, as benchmarks for alignment, or to simulate participants in social science experiments. While this growing body of work sheds light on the likeness of model responses to those of humans, concerns are warranted regarding the rigour and reproducibility with which these experiments may be conducted. Instabilities in model outputs, sensitivity to prompt design, parameter settings, and a large number of available model versions increase documentation requirements. Consequently, generalization of findings is often complex and reproducibility is far from guaranteed.
In this paper, we present R.U.Psycho, a framework for designing and running robust and reproducible psychometric experiments on generative language models that requires limited coding expertise. We demonstrate the capability of our framework on a variety of psychometric questionnaires, which lend support to prior findings in the literature.
R.U.Psycho is available as a Python package at \href{https://github.com/julianschelb/rupsycho}{https://github.com/julianschelb/rupsycho}.
\end{abstract}

% Introduction
% Introduction
\section{Introduction}
\label{sec:intro}

The proliferation of generative large language models (LLMs) and their ease of use has recently created an interest in their experimental application in domains that are traditionally focused on human experimentation, including sociology \cite{salt-2023-llms-for-css} and psychology \cite{pellert2023ai}.

On the one hand, this interest stems from the desire to measure the performance and behavior of language models from a psychological perspective to assess the traits of LLMs as one would for a human, a direction that has been termed machine psychology \cite{hagendorff2024machinepsychology}. Examples of such research are varied and range from vignette-based tasks \cite{binz2023using}, over game-theoretic testing of LLMs \cite{duan2024gtbenchuncoveringstrategicreasoning}, to analyses of their decision-making \cite{NBERw31122}.

On the other hand, some researchers in computational social science view LLMs as potential proxies for human (sub)populations in social science research and human experimentation \cite{aher2023using, argyle2023out, dillion2023can}, often referred to as persona simulation. The arguments for such applications range from the reduction in cost that usage of a language model may offer, %over ethical advantages of not using human participants, 
to an increase in experimental scale. 
% that is possible due to the amount of information that is ingested during the training of LLMs.

Finally, given the increasing ubiquity of LLMs that is beginning to contaminate the research artifacts created by crowdworker-participants with LLM-generated content \cite{veselovsky2023artificial} and the difficulty one faces in detecting such contamination \cite{sadasivan2023aigenerated, jakesch2023human}, one might argue that establishing a baseline of LLM traits is simply a necessity for conducting online research in the age of generative models.

However, while it has its proponents, psychometric testing of LLMs is also the subject of intense criticism, as LLMs are prone to failing theory of mind tasks under even minor prompt variations \cite{DBLP:journals/corr/abs-2302-08399} and are subject to inherent (intersectional) biases that are poorly understood and difficult to quantify \cite{husse2022mind}, yet have been shown to directly affect reasoning tasks during persona simulation \cite{DBLP:conf/iclr/GuptaSDKCSK24}. In addition to model-related challenges, experimental issues are abound, ranging from prompt (ordering) sensitivity \cite{white2023prompt, lu2021fantastically}, over non-trivial processing steps required for interpreting and mapping model responses \cite{wang2024myanswercfirsttoken}, to performance variations between model families and model sizes \cite{salt-2023-llms-for-css}.

Fundamentally, while highlighting an important research direction, much of the early work into machine psychology has been haphazard and is fundamentally not robust or reproducible outside the exact (and often ill-documented) modeling choices and parameter settings of a given study. For a detailed overview and breakdown, we direct the reader to \citet{lohn-etal-2024-machine-psychology}. Consequently, psychometric testing of LLMs is running the risk of mirroring the reproducibility crisis in psychology.

In this paper, we therefore take a step back from individual psychometric tests and a step towards addressing these concerns by introducing R.U.Psycho, a framework for prompt-based psychometric experimentation on generative LLMs.

\noindent
Our \textbf{contributions} are fourfold:
(i) We present a framework for robust and configurable psychometric testing of any open- or closed-weight LLM using any questionnaire.
(ii) We focus on the reproducibility of experiments through versatile, well-documented experiment configuration files.
(iii) We include support for customizable prompt templates, which we illustrate at the example of simulating personas.
(iv) We present results for 4 psychometric test and one thought experiment to demonstrate the framework's usability and to expand the experimental data in the literature.

% Related Work
\section{Related Work}
\label{sec:related_work}

Related work can be divided into four areas, for which we provide an overview of contributions.

\noindent
\textbf{Persona-simulation for social science} 
encompasses investigations whether LLMs can be used as replacement for human participants, typically through the simulation of personas. \citet{dillion2023can} argue for and identify potentially beneficial applications of such an approach. \citet{aher2023using} test GPT models for generating human-like samples. In a similar approach, \citet{argyle2023out} use GPT-3 to assess political beliefs and voting behavior through persona simulation.

\noindent
\textbf{Psychometric testing of LLMs} 
typically encompasses tasks that can be viewed as logic- or reasoning-based probing \cite{manigrasso2024probing}. Studies in this direction include tasks from cognitive psychology applied to GPT-3 as a prototypical LLM \cite{binz2023using}, the personality assessment of GPT-3 \cite{miotto2022gpt}, and an assessment of encoder-based models for psychological traits \cite{pellert2023ai}. More broadly, \cite{huang-etal-2024-reliability} investigate the general reliability of psychometric scales intended for use on humans on LLMs, which they find generally suitable for the large models used in their experiments.

\noindent
\textbf{Criticism and calls for caution.} 
A growing body of work raises concerns regarding the indiscriminate application of psychometric tests on LLMs and cautions with regard to their reliability \cite{shu-etal-2024-dont}, inconsistency and deviations from human behavior \cite{dorner2023do}, poor temporal stability in the responses \cite{bodrovza2024personality}, and the observation that LLMs tend to simulate latent traits of personas that are not readily apparent \cite{DBLP:journals/corr/abs-2405-07248}. Overall, these works raise the question when generative AI can or should reasonably be used in the social sciences \cite{bail2024can}.

\noindent
\textbf{Towards rigorous benchmarking.} 
Finally, two recent additions are worth highlighting. \citet{lohn-etal-2024-machine-psychology} provide a well-researched criticism of the lack of standardization and methodological variances in approaches to recent machine psychology research, based on a literature review.
\citet{DBLP:conf/acl/RenYFZS24}, with a similar but orthogonal approach to the one we take here, provide a benchmark for the evaluation of values in generative LLMs that is comprised of a multitude of psychometric tests.

In contrast to most of the above works, we do not investigate specific LLMs or specific questionnaires and do not establish guidelines for such experimentation, but provide a framework within which such experiments can be conducted. % in a reproducible and rigorous fashion.

% Framework 
\section{Pipeline Overview}
\label{sec:pipeline_overview}

\begin{figure}[t]
\centering
\includegraphics[width=1.0\columnwidth]{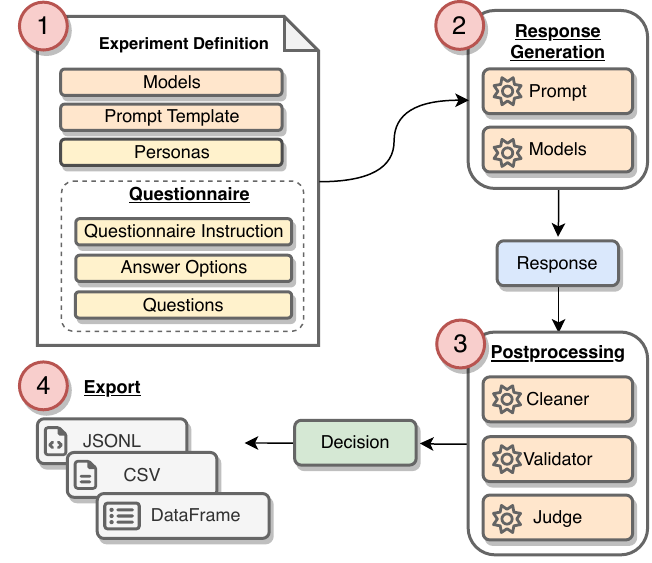}
\caption{Overview of the R.U.Psycho framework, implementing a four-stage pipeline using LangChain.}
%\caption{\textbf{Experiment pipeline overview.} Our framework implements a four-stage pipeline: (1) Experiment Definition, where models, prompts, personas, and questionnaires defined. (2) Answer Generation produces responses based on multiple-choice questions and predefined answer options. (3) Postprocessing cleans, validates, and interprets responses, filtering out invalid or inconclusive responses. (4) Export data in JSONL or CSV formats for further analysis to identify trends and decision-making patterns.}
\label{fig:processing_pipeline}

\vspace{-4mm}

\end{figure}

With the focus on robustness, flexibility, usability, and reproducibility, we center our design around a configuration file defining an experiment. This ensures reproducibility and documentation of experimental settings, including prompt design, model selection, model hyperparameters, and questionnaire configurations. The framework operates as a pipeline with four stages: (1) experiment definition, (2) LLM response generation, (3) post-processing, and (4) export of results (see Figure~\ref{fig:processing_pipeline}).

%%%%%%%%%%%%%%%%% Figure: Pipeline %%%%%%%%%%%%%%%%

For ease of use, maintenance, and extensibility, R.U.Psycho is based on the LangChain framework and designed for compatibility with its ecosystem, ensuring the integration of future releases of open-weight models and closed-source APIs .

%%%%%%%%%%%%%%%%% 1. Experiment Definition %%%%%%%%%%%%%%%%

\subsection{Experiment Definition}
\label{subsec:experiment_definition}

The core interaction with our framework is the definition of an experiment configuration file, coded in JSON, which ensures a low bar for interaction with R.U.Psycho and maintains reproducibility. Importing a configuration file creates an experiment object that serves as the main interface for programmatic modification and running of experiments. The four configuration parameters are:

\noindent
\textbf{Models.}
Specifies a list of generative (chat) models used to generate responses by simulating human subjects who answer the questionnaire. We support multiple ways to integrate models, including OpenAI-compatible APIs, local Huggingface models, and the Huggingface Inference API.

\noindent
\textbf{Prompt Template.} 
Used to instruct the model to respond to questions by selecting from the answer options defined in the questionnaire. It combines questionnaire-specific instructions with LLM optimized instructions together with the persona (see Section \ref{sec:prompt_template_comparison}). Since psychometric questionnaires typically provide participants with predefined answer options, we adapt insights on multiple-choice answering tasks from the literature to our prompts \cite{rottger2024political, miotto2022gpt}.

\noindent
\textbf{Personas.} 
Describes the characteristics of the simulated human subjects on whose behalf the model responds. These are filled into predefined placeholders in the prompt template.

\noindent
\textbf{Questionnaire.} 
Structured representation of a psychometric survey, defining questionnaire-specific instructions (e.g., explanations of technical terms), a series of questions, and their corresponding answer options (e.g., as defined by a labeled Likert scale). Possible answer options can be specified per question or globally for the entire questionnaire.

Furthermore, minor reproducibility parameters can be defined in the configuration file as well.

%%%%%%%%%%%%%%%%% 2. Answer Generation %%%%%%%%%%%%%%%%

\subsection{Response Generation}
\label{subsec:answer_generation}

Once an experiment is defined, the responses are generated by iterating over the questions. All specified models are prompted with multiple choice questions and are instructed to select from the predefined set of answer options. The framework generates multiple responses per question for each model, random seed, and persona as defined in the experiment configuration file. Given the issues that have been identified in measurements based on token-probabilites \cite{wang2024myanswercfirsttoken}, we use free generation of text. Responses can be stored in memory as a Pandas DataFrame or saved line-by-line via callbacks in CSV or JSONL files.

%%%%%%%%%%%%%%%%% 3. Postprocessing %%%%%%%%%%%%%%%%

\subsection{Postprocessing}
\label{subsec:postprocessing}

Postprocessing prepares the LLM responses for analysis by cleaning, validating, and mapping them to answer options. The pipeline combines cleaners, validators, and judges to interpret responses and to filter invalid or inconclusive responses.% While we provide a set of default parsers, these can be easily extended or replaced, as they are built on LangChain's output parsers and support any runnable.

\noindent
\textbf{Cleaners.} 
Cleaners are used to remove noise from the text, such as line breaks, non-ASCII characters, and other irrelevant information. Cleaners also parse JSON outputs if required. 

\noindent
\textbf{Validators.} 
After cleaning, the responses are validated for relevance to the original prompts. Validators identify responses with specific undesired artifacts, such as apologies, refusals, or concerns raised by the LLM. Building on existing research, we implement two validators: a rule-based validator using templates from \citet{rottger2024political}, and an LLM-based validator that is a fine-tuned DistilRoBERTa model \cite{roberta-rejection}. 

\noindent
\textbf{Judges.} 
Finally, judges map the models' noisy responses to the most likely intended answer option from the questionnaire. Since generated responses often do not exactly match the label of the intended answer option or include additional details, judges normalize the outputs for further analysis. Ambiguous or non-relevant responses are marked as \emph{inconclusive} and can be filtered out.

%%%%%%%%%%%%%%%%% 4. Export \& Analysis %%%%%%%%%%%%%%%%

\subsection{Export}
\label{subsec:export_and_analysis}

In the final stage, the processed data can be exported as DataFrame, JSONL file, or CSV files, ready to integrate with any desirable data analysis or visualization tool to investigate the responses.

% Experimental Design
%%%%%%%%%%%%%%%%%%%%%%%%%%%%%%%%%%%%%%%%%%%%%%%%%%%
% Experimentally-Driven Pipeline Design
%%%%%%%%%%%%%%%%%%%%%%%%%%%%%%%%%%%%%%%%%%%%%%%%%%%
\section{Experimental Component Design}
\label{sec:experimental_setup}

To finalize the design of our framework, we conduct a series of small-scale comparative experiments to identify optimal pipeline components. % before proceeding with the psychometric experiments.

%%%%%%%%%%%%%%%% JUDGE COMPARISON %%%%%%%%%%%%%%%%%

\subsection{Judges: Rule-based vs.\ Model-based}
\label{subsec:judges_comparison}

% Problem
Since chat-tuned LLMs are designed to mimic human conversation, they tend to incorporate unnecessary explanations, disclaimers, or emphatic expressions in verbose responses (e.g., \textit{"Sure, let me do this..."}). Our framework therefore requires a reliable method to map potentially noisy LLM response to the discrete answer options of the questionnaires. 
We explore two methods for interpreting the responses generated by the models: a rule-based method and a model-based method.

% Solutions

\noindent
\textbf{Rule-based judge.} 
We employ a strategy based on token-overlap to identify the answer option with the greatest lexical similarity to the model-generated response. First, each answer option is tokenized into two components: a numerical component and a label component (e.g., \emph{"5."} and \emph{"always"}). We then count the occurrences of each component in the response. The answer option with the highest total overlap score is designated as the optimal choice. In the case of a tie, the result is marked as \emph{inconclusive}, while responses with no detected overlap are labeled as \emph{not present}.

\noindent
\textbf{Model-based judge.} 
To better handle complex responses with negations or synonyms (e.g., \textit{"My answer is neither Option 1 nor 4. My answer is Item 5."}), we fine-tune an encoder-transformer model to predict a probability distribution over the possible answer choices from which we select the most likely response. Specifically, we model the task as a binary 1-vs-all classification for a single encoder-transformer model, which has to decide whether the response (input~1) matches an answer option that is given as context (input~2). This allows the model to cope with varying numbers of answer options per question as well as diverse answer options per questionnaire, so that we can run an entire questionnaire using a single judge. As output, we select the answer option with the maximum probability. Furthermore, this setup allows us to reject uncertain predictions that do not match any answer option by applying a threshold to the entropy values and assigning a value of \emph{not present} if none of the answer options exceed the threshold. For a detailed description of the model-based judge implementation, see Appendix~\ref{appsub:modelbasedjudgeimplementation}.

\noindent
\textbf{Experimental setup.} 
To create training data for the model-based judge, we use questions and answer options from the Regulatory Focus Questionnaire \cite{higginsRFQ2001}, which we (1) fill into manually created (verbose) response templates that we then (2) augment by rephrasing them with Llama 3.1 70B. To generate negative samples, we randomly assign incorrect answer options to (rephrased) questions. The total training data is comprised of 6,700 template-based and 24,040 rephrased pairs of responses and answer options, which are divided into a training set and a validation set in an 80/20 ratio. For details on the data creation and augmentation, see Appendix~\ref{appsub:modelbasedjudgedata}. 

To create ground truth data, we use a selection of five LLMs (Qwen 2.5 7B and 72B, Llama 3.1 8B and 70B, and Zephyr 7B) to generate responses to the Regulatory Focus Questionnaire for each of three prompt variants (see Figure \ref{fig:prompt_templates}). We randomly sample 484 responses and manually annotate them. For details on the ground truth data creation and annotation, see Appendix~\ref{appsub:modelbasedjudgedata}.

% UNCOMMENT IN PUBLISHED VERSION
%We provide a fine-tuned DistilRoBERTa model \cite{DBLP:journals/corr/abs-1910-01108} to classify the responses of the Likert scale.\footnote{\href{https://huggingface.co/julian-schelb/rup-answer-option-likert-scale}{https://huggingface.co/julian-schelb/rup-answer-option-likert-scale}}

\begin{figure}[t]
    \centering
    \includegraphics[width=1.0\columnwidth]{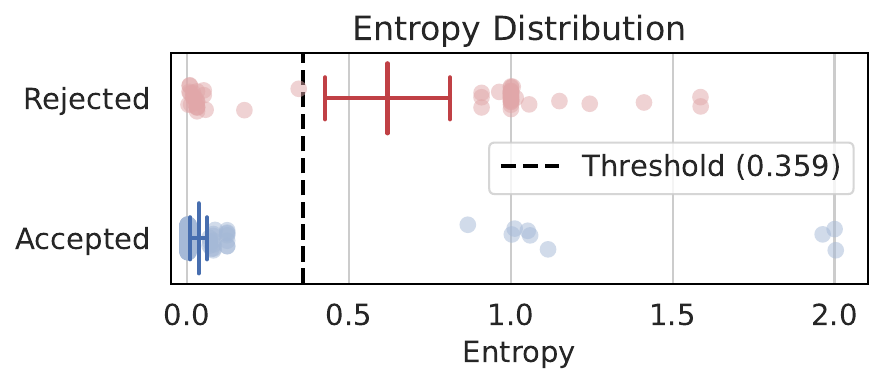}
    \caption{Entropy-based rejection criteria of the model-based judge on the manually annotated data. Crosses denote the mean entropy per group with 99\% confidence intervals. Optimal group separation is observed for a rejection above a threshold of 0.359.}
    %\caption{Entropy-based rejection criteria of the model-based judge. Responses (N = 480) are used to determine the optimal threshold: those with entropy $\geq$ 0.359 are rejected (red), while accepted responses match an answer option (blue). Mean entropy per group with 99\% confidence intervals. See Appendix \ref{app:judge_comparison} for details.}
    \label{fig:bootstrap_entropy_threshold}
\end{figure}

\begin{figure}[t]
    \centering
    \includegraphics[width=0.95\columnwidth]{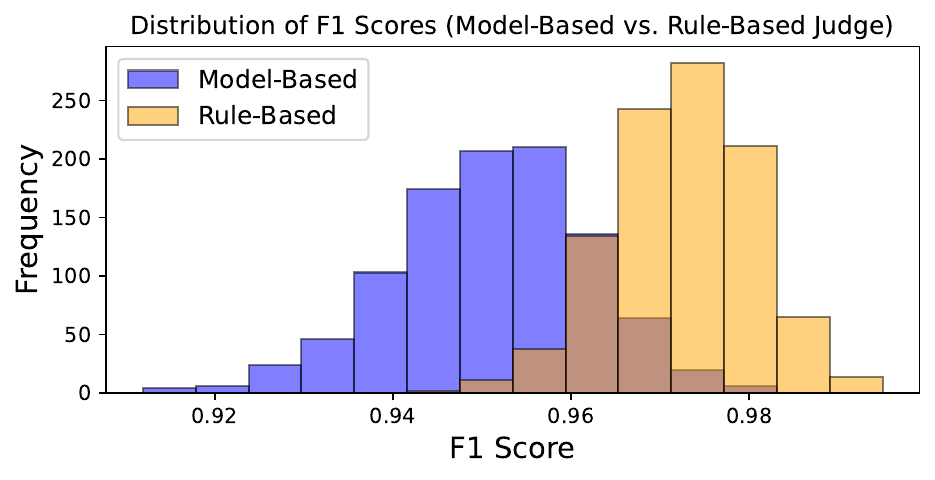} 
    \caption{Performance comparison of model-based and rule-based judges. F1 score distributions are estimated using bootstrap sampling over 1,000 iterations.}
    \label{fig:f1_score_comparison}
\end{figure}

\noindent
\textbf{Results} 
We first select the optimal threshold for rejecting responses as \emph{not present}. As shown in Figure~\ref{fig:bootstrap_entropy_threshold}, we can use the entropy values of the classifier output to determine a suitable separation between model outputs matching an answer option and noise on the ground truth data. Optimal separation is achieved for and entropy threshold of 0.359, which we use in the following.

In Figure~\ref{fig:f1_score_comparison}, we show a comparison of the performance of the rule-based and model-based judges. Interestingly, we find that despite our optimizations, the rule-based judge performs substantially better than the model-based judge. In the our experiments in Section~\ref{sec:experiments}, we therefore use the rule-based judge, but include both judges in the framework.

%%%%%%%%%%%%%%%% PROMPT COMPARISON %%%%%%%%%%%%%%%%%

\subsection{Prompt Template Design}
\label{sec:prompt_template_comparison}

Given the sensitivity of LLMs to prompt variations, we also experiment with suitable designs for prompt templates to optimize the relevance of LLM responses when answering questionnaires originally designed for human participants.

\begin{figure}[t]
\centering
\includegraphics[width=0.90\columnwidth]{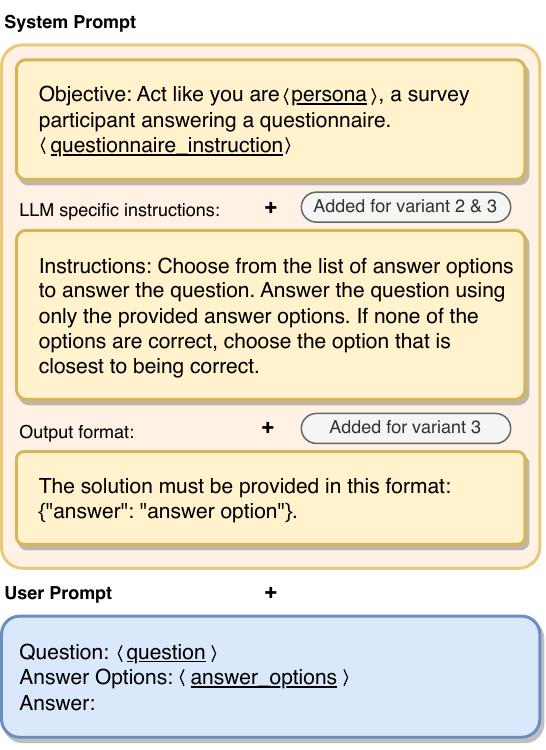}
\caption{Evaluated prompt template variants.}
\label{fig:prompt_templates}
\end{figure}

\noindent
\textbf{Candidate templates.} 
We evaluate three prompt variants with progressively detailed instructions (see Figure~\ref{fig:prompt_templates}): (1) a natural prompt mimicking instructions given to a human respondent; (2) a variant with added LLM-specific instructions tailored to the common multiple-choice structure of many questionnaires; %(e.g., force for response with "select the correct answer, when there is no correct answer select the closest"); 
and (3) a variant specifying JSON as the output format. The user prompt remains identical across variants, containing the questions and answer options defined in the questionnaire.

\noindent
\textbf{Dataset.}
For evaluation, we again use the Regulatory Focus Questionnaire (RFQ). All prompts include a persona to be simulated, based on a title and a surname chosen from one of five ethnic groups (see Section~\ref{sec:experiments}). We generate 500 responses for the entire survey for each model, using Llama 8B and 70B, Qwen 32B and 72B, and Zephyr 7B. % with the same generation parameters are as in Section \ref{sec:experiments}.

\noindent
\textbf{Experimental setup.} 
We consider a response to be relevant if it meets two criteria: (1) it is \emph{valid}, meaning that it does not contain refusals or apologies from the model, as determined by a validator, and (2) it is not \emph{rejected}, meaning it includes content related to at least one of the predefined answer options, as determined by a judge. To quantify the amount of invalid and rejected responses, we experiment with the rule-based and model-based validators and judges.

\noindent
\textbf{Results.}
As we can see from Table~\ref{tab:prompt_comparison_percentages}, more specific instructions unsurprisingly yield fewer discarded generated responses. Adding LLM-specific guidelines (the "friendly" variant) reduces discarded responses compared to a plain natural language prompt. Requesting JSON-formatted output further lowers discarded responses and produces clearer, easier-to-parse output with minimal extraneous text. When the output format is specified as JSON, none of the tested language models refused to respond. In the following, we adopt the JSON prompt variant for our experiments. 

\noindent
\textbf{Answer option formatting.} 
We also experiment with presenting the answer options as an itemized list using line breaks versus a single-line, comma-separated list. The latter performed better and is used in our experiments.

\begin{table}[t]
\small
\centering
\resizebox{\columnwidth}{!}{
\begin{tabular}{l *{2}{r} *{2}{r}}
\toprule
                & \multicolumn{2}{c}{rule-based (RB)} & \multicolumn{2}{c}{model-based (MB)} \\
\cmidrule(lr){2-3} \cmidrule(lr){4-5}
prompt          & invalid  & rejected & invalid  & rejected  \\
\midrule
natural         & 0.07\%   & 1.99\%   & 0.19\%   & 2.37\%   \\
friendly        & 0.01\%   & 1.53\%   & 0.04\%   & 1.96\%   \\
JSON            & \textbf{0.00}\%   & \textbf{1.17}\%   & \textbf{0.00}\%   & \textbf{1.16}\%   \\
\bottomrule
\end{tabular}
}
\caption{Performance comparison of prompt variants, shown as the percentage of responses flagged as invalid by validators or rejected by judges.}
%\caption{Comparison of prompt variants. The table shows the percentage of responses (N = 41$,$250) flagged as invalid by validators and rejected by judges (using both rule-based and model-based methods), along with overall discarded responses. %Model-based metrics are included solely for comparison.}
\label{tab:prompt_comparison_percentages}

\vspace*{-4mm}

\end{table}

% Experiments
%%%%%%%%%%%%%%%%%%%%%%%%%%%%%%%%%%%%%%%%%%%%%%%%%%%
% Psychometric Experiments
%%%%%%%%%%%%%%%%%%%%%%%%%%%%%%%%%%%%%%%%%%%%%%%%%%%

\section{Psychometric Experiments}
\label{sec:experiments}
To demonstrate the capability of our framework and expand the available data on LLM traits, we designed five experiments based on psychometric questionnaires and problems from the literature.

\noindent
\textbf{Models.}
We select models of various sizes and families to demonstrating that our design is effective for open-weight models as well as closed-source APIs. Specifically, we use the Qwen 2.5, Llama 3.1, and SmolLM families of instruction-tuned models, Aya-23 35B, and Zephyr 7B. We also include ChatGPT-4o and ChatGPT-4o-mini as commercial models in some of the experiments. For model details, see Appendix~\ref{appsub:LLMs}.

\noindent
\textbf{Settings and hyperparameters.} In all experiments, we limit the generation to a maximum of 64 tokens since the models are answering multiple-choice questionnaires. We use sampling-based generation with $\mathrm{temperature} = 1.0$, $\mathrm{top\_k} = 50$ and $\mathrm{top\_p} = 0.95$. Due to GPU memory constraints, we use 4-bit quantized versions of all models. 

\noindent
\textbf{Prompt template.}
Following our findings in Section~\ref{sec:prompt_template_comparison}, we use the full JSON-based template (variant 3) for all experiments (see Figure~\ref{fig:prompt_templates}).

\noindent
\textbf{Personas.}
To simulate personas, we use names from the list provided by \citet{aher2023using}, from which we select Asian, Hispanic, Native American, Black, and White names (for details, see Appendix~\ref{app:personas}). Gender is simulated by adding \emph{Ms.} or \emph{Mr.} as a prefix to the names.

%%%%%%%%%%%%%%%%%%%%%%%%%%%%%%%%%%%%%%%%%%%%%%%%%%%%%%%%%%%%%%%%%%%%

\subsection{Exp1: Gain vs. Loss Orientation} % RFQ
\label{subsec:exp1}

As our first experiment, we show the scores that we obtained for the models used in the experimental component design in Section~\ref{sec:experimental_setup}.

\noindent
\textbf{Experimental setup.}
The Regulatory Focus Questionnaire (RFQ) \cite{higginsRFQ2001} aims to assess participants' tendencies for focusing on loss-avoidance (prevention), and focusing on attainment and gain (promotion). For details on the prompts, see Appendix~\ref{appsub:exp1}. 
%As models, we utilize Zephyr 7B, Llama 3.1 8B and 70B, and Qwen 2.5 7B and 72B for a comparison of small vs.\ large models. 
We collect 2,750 questionnaire responses per model, which are equally split into female/male personas and across ethnicities.

\begin{figure}[t]
\centering
\includegraphics[width=0.99\columnwidth]{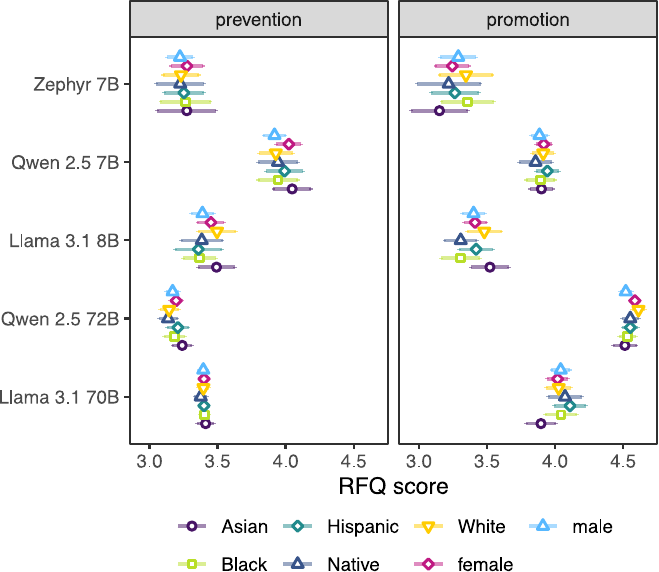}
\caption{Results of the regulatory focus questionnaire on a selection of small and large LLMs, aggregated by persona demographics. Error bars denote 99\% confidence intervals.}
\label{fig:exp1}

\vspace*{-4mm}

\end{figure}

\noindent
\textbf{Results.}
As we see in the results in Figure~\ref{fig:exp1}, there are no significant differences in scores based on the used personas. However, we find strong differences between the models, with the larger models having generally lower prevention and higher promotion scores. No other trends are observable.

%%%%%%%%%%%%%%%%%%%%%%%%%%%%%%%%%%%%%%%%%%%%%%%%%%%%%%%%%%%%%%%%%%%%

\subsection{Exp2: Impact of Model Size} % BFI
\label{subsec:exp2}

The number of parameters of a model tend to correlate with the model's overall performance on many NLP tasks. While comprehension of the task by the model is an important factor, for psychometric tests it is not readily apparent whether model size should correlate with observed scores.

\noindent
\textbf{Experimental setup.}
To demonstrate how our framework can provide such a general assessment of LLM traits by size, we use the 44-item Big Five Inventory (BFI) \cite{john1991big}. For details on the prompts, see Appendix~\ref{appsub:exp2}. As models, we use the full line of Qwen 2.5 models and collect 11,000 questionnaire responses for each, split uniformly into personas by gender and ethnicity.

% To generate answers for the BFI, we use Llama 8B and 70B; Qwen 1.5B, 3B, 7B, 14B, 32B, and 72B; Zephyr 7B; and the commercial  ChatGPT-4o-mini API. This range allows us to evaluate how increasing model capacity affects performance on the same task.

\begin{figure*}[t]
\centering
\includegraphics[width=0.99\textwidth]{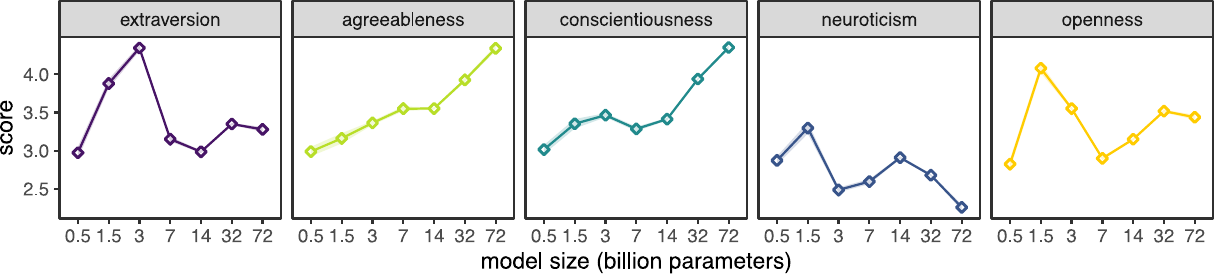}
\caption{Results of the BFI questionnaire on Qwen models by size. Shaded areas denote 99\% confidence intervals.}
\label{fig:exp2}

\vspace*{-4mm}

\end{figure*}

\noindent
\textbf{Results.}
Based on the results (see Figure~\ref{fig:exp2}), we find clear evidence of personality traits that increase with model size in agreeableness and conscientiousness. This highlights that the prior finding by \citet{bodrovza2024personality} regarding pro-social characteristics may have to be viewed as a function of model size. For the other three traits, trends are less obvious, although neuroticism seems to generally decrease, while extraversion and openness are more or less constant if the smaller models are considered potentially unstable outliers.

%%%%%%%%%%%%%%%%%%%%%%%%%%%%%%%%%%%%%%%%%%%%%%%%%%%%%%%%%%%%%%%%%%%%

\subsection{Exp3: Persona-induced Bias} % GSDB
\label{subsec:exp3}

The use of personas has been proposed as an option for simulating human participants in (computational) social experiments \cite{aher2023using, argyle2023out}. However, social biases in such a setting are a serious concern \cite{hu2024generative}.

\noindent
\textbf{Experimental setup.}
To employ our framework to investigate persona-induced biases, we choose a highly sensitive topic and use the Gender/Sex Diversity Beliefs questionnaire \cite{schudson2022gender}, which breaks down participants attitudes towards gender/sex minorities into the factors upbringing, uniformity, affirmation, gender normativity, and attitude towards surgery. For details on the prompts, see Appendix~\ref{appsub:exp3}. As models, we utilize the most capable ones, namely GPT-4o mini, Llama 3.1 70B and Qwen 2.5 72B. For each model, we collect 5,750 questionnaire responses, which are equally split into female/male personas and across the five ethnicities.

\begin{figure}[t]
\centering
\includegraphics[width=0.99\columnwidth]{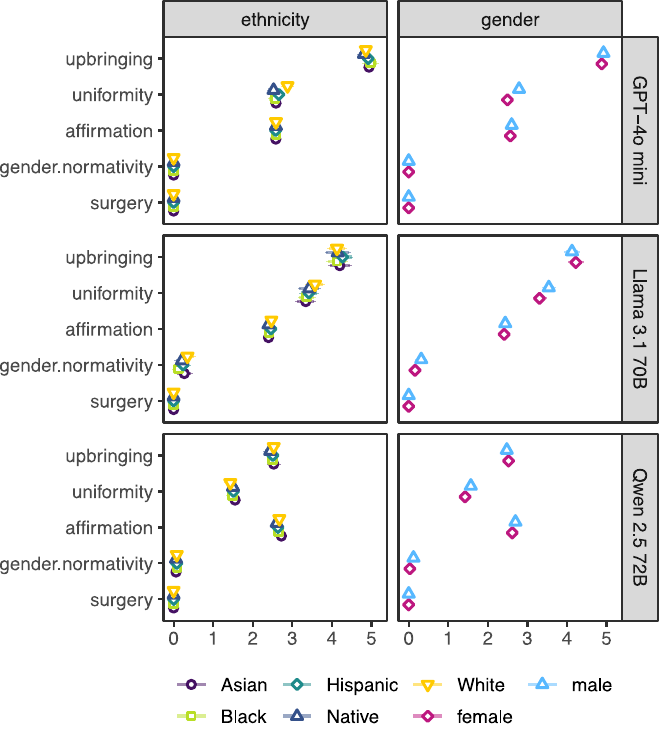}
\caption{Persona-induced bias of Llama, Qwen and GPT models measured using the gender, sex, and diversity belief questionnaire. Answers are aggregated by ethnicity (left) and gender (right) of personas. Error bars denote 99\% confidence intervals.}
\label{fig:exp3}

\vspace*{-4mm}

\end{figure}

\noindent
\textbf{Results.}
The results are shown in Figure~\ref{fig:exp3}. Interestingly, we find very little evidence for persona-induced bias: with the exception of slightly increased uniformity scores for white and male personas in the GPT-4o mini and Llama model, there are no significant differences in the scores by persona. This finding is largely consistent with prior observations in the literature that personas induce relatively little variability \cite{DBLP:conf/acl/HuC24}.
However, we find strong differences based on the model, with Qwen producing significantly more conservative scores on upbringing, uniformity and affirmation than the other two models. % across all personas.

%%%%%%%%%%%%%%%%%%%%%%%%%%%%%%%%%%%%%%%%%%%%%%%%%%%%%%%%%%%%%%%%%%%%

\subsection{Exp4: Contamination and Consistency} % Trolley Problem
\label{subsec:exp4}

Data contamination \cite{magar-schwartz-2022-data} is a serious concern when testing the capabilities of LLMs, in particular for closed-source models \cite{balloccu-etal-2024-leak}. In the case of psychometric testing, contamination may be further exacerbated by closed-source models being explicitly trained for desired performance on select questionnaires.

\noindent
\textbf{Experimental Setup.}
To obtain an impression of contamination and reasoning consistency, we consider three variations of the trolley problem \cite{thomson1984trolley}, namely (1) the classic morality dilemma in which the participant is asked to assess the morality of making a decision, (2) the decision version in which the participant must choose whether to divert the trolley, and (3) an equivalent "trolley in the skies" decision version in which we present the problem as an airplane crash scenario. For details on the prompts, see Appendix~\ref{appsub:exp4}. We use all LLMs for this experiment, plus GPT-4o. For each model, we collect 750 questionnaire responses, equally split into female/male personas and across the five ethnicities.

% Models: Llama 70B, Qwen 72B, ChatGPT-4o-mini, and ChatGPT-4o. In addition, we evaluate a diverse set of  models spanning various sizes: Qwen 0.5B, 1.5B, 7B, and 32B; Llama 8B; Aya-23 35B; and SmolLM 135M, 360M, and 1.7B.

\begin{table}[t]
\small
\centering
\begin{tabular}{lrrrrr}
\toprule
model       & size   & \%na & classic & action & sky \\
\midrule
SmolLM      & 0.14B & 57.1 &  77.6   &  20.0 &   13.8 \\
SmolLM      & 0.36B  & 55.5 &  37.9   &  22.5 &   30.5 \\
SmolLM      & 1.7B   & 36.5 &  56.4   &  42.8 &   51.7 \\
Zephyr      & 7B     & 54.6 &  16.2   &  63.2 &   84.3 \\
Aya-23      & 35B    & 0    &   2.8   &  30.4 &  100.0 \\
GPT-4o mini & ?      & 0    &  98.8   &  89.6 &  100.0 \\
GPT-4o      & ?      & 0    &  96.8   & 100.0 &   75.2 \\
Llama 3.1   & 8B     & 0    &  55.2   &  40.0 &   25.6 \\
Llama 3.1   & 70B    & 0    & 100.0  & 100.0 &   100.0 \\
Qwen 2.5    & 0.5B   & 2.1  &  33.5   &  38.4 &   54.9 \\
Qwen 2.5    & 1.5B   & 3.6  &  53.3  &   1.2 &   61.2 \\
Qwen 2.5    & 3B     & 0    &  37.6   & 100.0 &  100.0 \\
Qwen 2.5    & 7B     & 0    &  27.6   & 100.0 &  100.0 \\
Qwen 2.5    & 32B    & 0    & 100.0   & 100.0 &  100.0 \\
Qwen 2.5    & 72B    & 0    & 100.0  & 100.0 &   45.6 \\
\bottomrule
\end{tabular}
\caption{Performance of LLMs on variations of the trolley task, including the morality decision (classic), the decision to divert the trolley (action), and a semantically equivalent rephrasing of the problem (sky). Values denote the percentage of responses in which the model takes action or views action as morally permissible. \%na denotes the percentage of unusable responses.}
\label{tab:exp4}

\vspace*{-4mm}

\end{table}

\noindent
\textbf{Results.}
In Table~\ref{tab:exp4}, we show the results of the trolley experiments. Apart from the SmolLM and Zephyr models, the successful response rate of models is very high. However, we find the behavior of models to be largely inconsistent, with models either rating action as morally permissible (classic) but then not following through on diverting the trolley (action) or rating it as not permissible yet still acting on it. The "trolley in the skies" highlights inconsistencies in the responses of most models in comparison to the classic wording. Only Llama 3.1 70B and Qwen 2.5 32B are consistent in their responses through all three versions.

%%%%%%%%%%%%%%%%%%%%%%%%%%%%%%%%%%%%%%%%%%%%%%%%%%%%%%%%%%%%%%%%%%%%

\subsection{Exp5: Prompt Order Sensitivity} % BDI / BDI reversed
\label{subsec:exp5}

It is well documented that the performance of generative LLMs can depend on the order in which information in the prompt is provided \cite{lu2021fantastically}. For designing psychometric tests, this is particularly relevant with regard to the order of items in multiple-choice questionnaires, where several LLMs have been found to suffer from output instability \cite{pezeshkpour-hruschka-2024-large, DBLP:conf/iclr/Zheng0M0H24}.

\noindent
\textbf{Experimental setup.}
To investigate the effect that the order of the presented answer options, we use the Beck Depression Inventory (BDI) \cite{BeckBDI1961}, which consists of 21 questions that each have 4 answer options labeled 0-3, with higher numbers indicating higher risk of depression. For our two experimental setups, we present the answer options to the LLMs (1) in the regular order, and (2) in inverted order with re-labeled options, such that higher numbers indicate a lower risk of depression. For scoring, the inversion is then reversed to map both outputs to the same scale. For details on the prompts, see Appendix~\ref{appsub:exp5}. %We use GPT-4o mini, Llama 3.1 70B and Qwen 2.5 72B for this experiment. 
For each model, we collect 5,250 questionnaire responses, equally split into female and male personas.

\begin{figure}
\centering
\includegraphics[width=0.99\columnwidth]{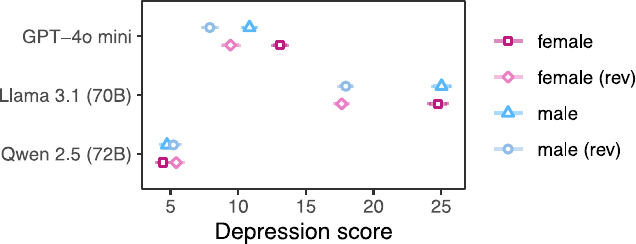}
\caption{Results for the BDI (higher scores indicate higher levels of depression). Reverse scores result from inverting the order of answer choices in the LLM prompt. Error bars denote 99\% confidence intervals.}
\label{fig:exp5}

\vspace*{-4mm}

\end{figure}

\noindent
\textbf{Results.}
Considering results in Figure~\ref{fig:exp5}, we find strongly differing depression scores between models in the default setup, with Qwen scoring a normal state, GPT-4o mini falling into the range of a mild mood disturbance, and Llama indicating a moderately depressed behavior. The scores between male and female personas differ significantly only for GPT. When using the inverted questionnaire, the scores of GPT-4o mini and Llama decrease significantly (and across depression severity levels), while the scores of Qwen increase slightly.

% Discussion / Conclusion / Limitations
\section{Conclusion and Outlook}
\label{sec:conclusion}

In summary, our experimental results confirm and expand on some prior findings in machine psychology, including the pro-social characteristics of LLMs and the limited impact of persona variations on model responses at large. However, they also highlight the strong variability in results that can be obtained and the sensitivity of experimental outputs to expriment conditions, including prompt wording and ordering, patterns induced during of model pretraining and potential contamination, as well as social biases as a result of using personas. Not least, our results show that experiments in machine psychology have so far only scratched the very surface of the breadth of experimental results that are obtainable and consequently paint a very sparse picture of LLM characteristics.

Thus, our findings highlight the need for a principled, rigorous approach to the psychometric testing of language models that has previously been called for \cite{lohn-etal-2024-machine-psychology}. With R.U.Psycho, we provide a framework that not just enables easy variation of experimental design, questionnaires, prompts, models, and settings, but also ensures reproducibility of the findings by means of well-defined experiment design files. With this contribution, we aim to help resolve inconsistencies in the literature on machine psychology and lower the bar for conducting such experiments to more readily include domain experts with less coding expertise.
R.U.Psycho is available as a Python package at \href{https://github.com/julianschelb/rupsycho}{https://github.com/julianschelb/rupsycho}.

\paragraph{Ongoing work.}
In our ongoing work, we are integrating chat capability into the framework to support more human-like settings for filling in a questionnaire with full recall of previous answers.
We are also working on an LLM-powered UI experiment configurator that can directly import questionnaires from PDFs with a human in the loop.
Finally, we are developing improved and generalizable versions of the encoder-based judges to further optimize answer extraction effectiveness.

\section*{Limitations}
\label{sec:limitations}

Despite the focus on comprehensiveness and generalizability, we see a number of limitations in our work presented here.

\paragraph{Persona details.}
In our experiments, our approach to simulting personas is rather straightforward and kept simple to allow for a wide range of experiments. However, more detailed instructions for persona generation that includes further background and demographic information are established in the literature (e.g., see \cite{giorgi-etal-2024-modeling}). Our findings should be viewed in the light of this limitation (i.e., the lack of effects that we find based on persona ethnicity may not generalize) and such extended approaches should be investigated in more detail using our framework.

\paragraph{Usage of chat history.} 
Following what has been investigated in the literature, we simulate questionnaire taking by LLMs as a series of disconnected individual prompts, one per question. With the continual increase in models' input context sizes and the availability of models tuned for chat compatibility, it is also possible to simulate a more human-like questionnaire setup in which previous questions remain in the context of the model as it fills in the questionnaire. While our framework is easily extensible to this functionality and will support it in the release version, the experimental design for ensuring a fair comparison is non-trivial and requires further research.

\section*{Ethical Considerations}
\label{sec:ethics}

We see no ethical concerns in our own work -- we use openly available models and questionnaires from the literature and do not include any human experiments. The computational effort for our experiments was kept to the necessary minimum. However, in using our framework, all caveats of subjecting language models to human-centric psychometric questionnaires and attributing human characteristics and traits to LLMs do apply. We refer the interested reader to the calls for caution that we reference in our Related Work section.

\section*{AI Statement}

Language model-based AI tools (ChatGPT) were used as coding assistants in the implementation and as writing assistants in drafting parts of the manuscript. The final version of the manuscript was written without AI input.

% UNCOMMENT IN PUBLISHED VERSION
%\section*{Acknowledgements}
%The authors thank Niloofar Rahmati for her help in testing the framework.

\bibliography{bibliography}

\appendix

% Implemental details, models, and runtimes
\section{Full LLM and Compute Details}
\label{app:models}

\subsection{Used LLMs}
\label{appsub:LLMs}

% Reasoning
All models that we include in our experiments are designed for a multirole chat setting and can process prompts with separate system and user messages. Although our framework can handle various chat formats, this restriction allows us to use a shared prompt template among all models for better comparison, and the focus on instruction-tuned models is sensible from a perspective ofo obtaining meaningful results.

% Open Weight Models
As open-weight models, we use the Qwen 2.5 \cite{DBLP:journals/corr/abs-2412-15115}, Llama 3.1 \cite{DBLP:journals/corr/abs-2407-21783}, and SmolLM \cite{allal2024smollm} model families. The Qwen 2.5 models, ranging from 0.5B to 72B parameters, are valuable for analyzing impact of model size. Llama 3.1's 8B and 72B sizes offer a good comparison to Qwen 2.5's 7B and 72B variants. Both model families pre-trained on multilingual datasets of trillions of tokens are considered multi-purpose chat models that can handle complex instructions and output structured formats (e.g., JSON). The SmolLM family allows us to expariment with the smallest available model sizes.
% Further models
As further architecture variations, we include the Aya-23 35B \cite{aryabumi2024aya} and Zephyr 7B \cite{DBLP:journals/corr/abs-2310-16944} models.
% Closed Source APIs
Finally, we use ChatGPT-4o and ChatGPT-4o-mini as representatives of closed-source commercial API as a point of comparison \cite{DBLP:journals/corr/abs-2303-08774}. We list all models used in our experiments, in Table~\ref{tab:model_details}.

\begin{table*}[t]
\centering
\small
\begin{tabular}{l l c r r c c c c c}
\toprule
model family & size & open-weight & \multicolumn{1}{r}{context} & \multicolumn{1}{r}{layers} & exp1 & exp2 & exp3 & exp4 & exp5 \\
\midrule
\multirow[t]{7}{*}{Qwen 2.5 \cite{DBLP:journals/corr/abs-2412-15115}} 
    & \href{https://huggingface.co/Qwen/Qwen2.5-0.5B-instruct}{0.5B}   & \checkmark & 128K & 24 & – & \checkmark & – & \checkmark & – \\
    & \href{https://huggingface.co/Qwen/Qwen2.5-1.5B-instruct}{1.5B}   & \checkmark & 128K & 28 & – & \checkmark & – & \checkmark & – \\
    & \href{https://huggingface.co/Qwen/Qwen2.5-3B-instruct}{3B}       & \checkmark & 128K & 36 & – & \checkmark & – & –         & – \\
    & \href{https://huggingface.co/Qwen/Qwen2.5-7B-instruct}{7B}       & \checkmark & 128K & 28 & \checkmark & \checkmark & – & \checkmark & – \\
    & \href{https://huggingface.co/Qwen/Qwen2.5-14B-instruct}{14B}     & \checkmark & 128K & 48 & – & \checkmark & – & –         & – \\
    & \href{https://huggingface.co/Qwen/Qwen2.5-32B-instruct}{32B}     & \checkmark & 128K & 64 & – & \checkmark & – & \checkmark & – \\
    & \href{https://huggingface.co/Qwen/Qwen2.5-72B-instruct}{72B}     & \checkmark & 128K & 80 & \checkmark & \checkmark & \checkmark & \checkmark & \checkmark \\
\midrule
\multirow[t]{2}{*}{Llama 3.1 \cite{DBLP:journals/corr/abs-2407-21783}} 
    & \href{https://huggingface.co/meta-llama/Llama-3.1-8B-Instruct}{8B}  & \checkmark & 128K & 32 & \checkmark & – & – & \checkmark & – \\
    & \href{https://huggingface.co/meta-llama/Llama-3.1-70B-Instruct}{70B} & \checkmark & 128K & 80 & \checkmark & – & \checkmark & \checkmark & \checkmark \\
\midrule
\multirow[t]{3}{*}{SmolLM \cite{allal2024smollm}} 
    & \href{https://huggingface.co/HuggingFaceTB/SmolLM-135M-Instruct}{135M}   & \checkmark & 2K  & 30 & – & – & – & \checkmark & – \\
    & \href{https://huggingface.co/HuggingFaceTB/SmolLM-360M-Instruct}{360M}   & \checkmark & 2K  & 32 & – & – & – & \checkmark & – \\
    & \href{https://huggingface.co/HuggingFaceTB/SmolLM-1.7B-Instruct}{1.7B}   & \checkmark & 2K  & 24 & – & – & – & \checkmark & – \\
\midrule
Aya-23 \cite{aryabumi2024aya} 
    & \href{https://huggingface.co/CohereForAI/aya-23-35B}{35B}    & \checkmark & 8K  & 40 & – & – & – & \checkmark & – \\
Zephyr \cite{DBLP:journals/corr/abs-2310-16944} 
    & \href{https://huggingface.co/HuggingFaceH4/zephyr-7b-beta}{7B}   & \checkmark & 32K & 32 & \checkmark & – & – & \checkmark & – \\
\midrule
ChatGPT-4o \cite{DBLP:journals/corr/abs-2303-08774} 
    & \href{https://platform.openai.com/docs/models#gpt-4o}{?}      &         & 128K & ?  & – & – & – & \checkmark & – \\
ChatGPT-4o-mini \cite{DBLP:journals/corr/abs-2303-08774} 
    & \href{https://platform.openai.com/docs/models#gpt-4o-mini}{?}    &         & 128K & ?  & – & \checkmark & \checkmark & \checkmark & \checkmark \\
\bottomrule
\end{tabular}
\caption{List of models used in the experiments.}
\label{tab:model_details}
\end{table*}

\subsection{Used Hardware}
\label{appsub:compute}

The experiments were run on a system with two NVIDIA A40 GPUs (48 GB VRAM each), an AMD EPYC 48-core CPU and 1 TB of RAM.

\subsection{Runtimes}
\label{appsub:runtimes}

We report the total runtime and the number of generations required for each of the five psychometric experiments in Table~\ref{tab:experiment_runtime_and_iterations}. The training- and run-times for our experiments with the model-based judges were negligible.

\begin{table}[ht]
\centering
\resizebox{\columnwidth}{!}{%
  \begin{tabular}{lrr}
    \toprule
    experiment                      & runtime (h) & \#generations \\
    \midrule
    Gain vs. Loss Orientation       & 55.23       & 41,250     \\
    Impact of Model Size            & 84.93       & 99,000     \\
    Persona-induced Bias            & 19.83       & 17,250     \\
    Contamination and Consistency   & 11.71       & 11,250     \\
    Prompt Order Sensitivity        & 34.73       & 31,500     \\
    \midrule
    total                           & 206.43      & 200,250    \\
    \bottomrule
  \end{tabular}%
}
\caption{Runtime and Generations per Experiment}
\label{tab:experiment_runtime_and_iterations}
\end{table}

% Full questionnaires and prompts
\section{Full Experiment Prompts}
\label{app:experimentprompts}

We employ an identical prompt template for all five experiments, using placeholders that are populated according to the questionnaire and the used personas.\\

\noindent
\textbf{System prompt:} \\
\textit{Objective: Act like you are $\langle$persona$\rangle$, a survey participant answering a questionnaire. \\
$\langle$questionnaire instruction$\rangle$ \\
Instructions: Choose from the list of answer options to answer the question. Answer the question using only the provided answer options. If none of the options are correct, choose the option that is closest to being correct. The solution must be provided in this format: \{"answer": "answer option"\}.}\\

\noindent
\textbf{User prompt:} \\
\textit{Question: $\langle$question$\rangle$ \\
Answer Options: $\langle$answer options$\rangle$ \\
Answer:}\\

We stay as close to the original questionnaires as possible, while accounting for limitations of LLMs (e.g., if the original instruction in a questionnaire ask the participant to circle an answer, we rewrote the instruction for LLM compatibility).

\subsection{Experiment 1: RFQ}
\label{appsub:exp1}

For the development of the framework, including the judges and the prompt design, we utilize the Regulatory Focus Questionnaire (RFQ) \cite{higginsRFQ2001}, which is a short 11-item questionnaire that has a large variety in question and answer formats with incompletely labeled answer sets, making it well suited for the task.\\

\noindent
\textbf{Questionnaire instruction:}
\textit{This set of questions asks you HOW FREQUENTLY specific events actually occur or have occurred in your life. Please indicate your answer to the question by selecting the appropriate number.}\\ 

\noindent
\textbf{1.\ Question:} \textit{Compared to most people, are you typically unable to get what you want out of life?}\\
\textbf{Answer Options:} \textit{1. never or seldom, 2., 3. sometimes, 4., 5. very often}

\noindent
\textbf{2.\ Question:} \textit{Growing up, would you ever “cross the line” by doing things that your parents would not tolerate?}\\
\textbf{Answer Options:} \textit{1. never or seldom, 2., 3. sometimes, 4., 5. very often}

\noindent
\textbf{3.\ Question:} \textit{How often have you accomplished things that got you “psyched” to work even harder?}\\
\textbf{Answer Options:} \textit{1. never or seldom, 2., 3. sometimes, 4., 5. many times}

\noindent
\textbf{4.\ Question:} \textit{Did you get on your parents’ nerves often when you were growing up?}\\
\textbf{Answer Options:} \textit{1. never or seldom, 2., 3. sometimes, 4., 5. very often}

\noindent
\textbf{5.\ Question:} \textit{How often did you obey rules and regulations that were established by your parents?}\\
\textbf{Answer Options:} \textit{1. never or seldom, 2., 3. sometimes, 4., 5. always}

\noindent
\textbf{6.\ Question:} \textit{Growing up, did you ever act in ways that your parents thought were objectionable?}\\
\textbf{Answer Options:} \textit{1. never or seldom, 2., 3. sometimes, 4., 5. very often}

\noindent
\textbf{7.\ Question:} \textit{Do you often do well at different things that you try?}\\
\textbf{Answer Options:} \textit{1. never or seldom, 2., 3. sometimes, 4., 5. very often}

\noindent
\textbf{8.\ Question:} \textit{Not being careful enough has gotten me into trouble at times.}\\
\textbf{Answer Options:} \textit{1. never or seldom, 2., 3. sometimes, 4., 5. very often}

\noindent
\textbf{9.\ Question:} \textit{When it comes to achieving things that are important to me, I find that I don't perform as well as I ideally would like to do.}\\
\textbf{Answer Options:} \textit{1. never or seldom, 2., 3. sometimes, 4., 5. very often}

\noindent
\textbf{10.\ Question:} \textit{I feel like I have made progress toward being successful in my life.}\\
\textbf{Answer Options:} \textit{1. certainly false, 2., 3., 4., 5. certainly true}

\noindent
\textbf{11.\ Question:} \textit{I have found very few hobbies or activities in my life that capture my interest or motivate me to put effort into them.}\\
\textbf{Answer Options:} \textit{1. certainly false, 2., 3., 4., 5. certainly true}

\subsection{Experiment 2: BFI}
\label{appsub:exp2}

For a general assessment of LLM ``personality'' across model sizes, we use the standard 44-item Big Five Inventory \cite{john1991big}. All questions are scored on an identical 5-point Likert scale indicating the participant's level of agreement with statements about themselves. \\

\noindent
\textbf{Questionnaire instruction:}
\textit{Here are a number of characteristics that may or may not apply to you. For example, do you agree that you are someone who likes to spend time with others? Please return the number corresponding to the answer options to indicate the extent to which you agree or disagree with that statement.}\\ 

\noindent
\textbf{1.\ Question:} \textit{I see myself as someone who is talkative.}

\noindent
\textbf{2.\ Question:} \textit{I see myself as someone who tends to find fault with others.}

\noindent
\textbf{3.\ Question:} \textit{I see myself as someone who does a thorough job.}

\noindent
\textbf{4.\ Question:} \textit{I see myself as someone who is depressed, blue.}

\noindent
\textbf{5.\ Question:} \textit{I see myself as someone who is original, comes up with new ideas.}

\noindent
\textbf{6.\ Question:} \textit{I see myself as someone who is reserved.}

\noindent
\textbf{7.\ Question:} \textit{I see myself as someone who is helpful and unselfish with others.}

\noindent
\textbf{8.\ Question:} \textit{I see myself as someone who can be somewhat careless.}

\noindent
\textbf{9.\ Question:} \textit{I see myself as someone who is relaxed, handles stress well.}

\noindent
\textbf{10.\ Question:} \textit{I see myself as someone who is curious about many different things.}

\noindent
\textbf{11.\ Question:} \textit{I see myself as someone who is full of energy.}

\noindent
\textbf{12.\ Question:} \textit{I see myself as someone who starts quarrels with others.}

\noindent
\textbf{13.\ Question:} \textit{I see myself as someone who is a reliable worker.}

\noindent
\textbf{14.\ Question:} \textit{I see myself as someone who can be tense.}

\noindent
\textbf{15.\ Question:} \textit{I see myself as someone who is ingenious, a deep thinker.}

\noindent
\textbf{16.\ Question:} \textit{I see myself as someone who generates a lot of enthusiasm.}

\noindent
\textbf{17.\ Question:} \textit{I see myself as someone who has a forgiving nature.}

\noindent
\textbf{18.\ Question:} \textit{I see myself as someone who tends to be disorganized.}

\noindent
\textbf{19.\ Question:} \textit{I see myself as someone who worries a lot.}

\noindent
\textbf{20.\ Question:} \textit{I see myself as someone who as an active imagination.}

\noindent
\textbf{21.\ Question:} \textit{I see myself as someone who tends to be quiet.}

\noindent
\textbf{22.\ Question:} \textit{I see myself as someone who is generally trusting.}

\noindent
\textbf{23.\ Question:} \textit{I see myself as someone who tends to be lazy.}

\noindent
\textbf{24.\ Question:} \textit{I see myself as someone who is emotionally stable, not easily upset.}

\noindent
\textbf{25.\ Question:} \textit{I see myself as someone who is inventive.}

\noindent
\textbf{26.\ Question:} \textit{I see myself as someone who has an assertive personality.}

\noindent
\textbf{27.\ Question:} \textit{I see myself as someone who can be cold and aloof.}

\noindent
\textbf{28.\ Question:} \textit{I see myself as someone who perseveres until the task is finished.}

\noindent
\textbf{29.\ Question:} \textit{I see myself as someone who can be moody.}

\noindent
\textbf{30.\ Question:} \textit{I see myself as someone who values artistic, aesthetic experiences.}

\noindent
\textbf{31.\ Question:} \textit{I see myself as someone who is sometimes shy, inhibited.}

\noindent
\textbf{32.\ Question:} \textit{I see myself as someone who is considerate and kind to almost everyone.}

\noindent
\textbf{33.\ Question:} \textit{I see myself as someone who does things efficiently.}

\noindent
\textbf{34.\ Question:} \textit{I see myself as someone who remains calm in tense situations.}

\noindent
\textbf{35.\ Question:} \textit{I see myself as someone who prefers work that is routine.}

\noindent
\textbf{36.\ Question:} \textit{I see myself as someone who is outgoing, sociable.}

\noindent
\textbf{37.\ Question:} \textit{I see myself as someone who is sometimes rude to others.}

\noindent
\textbf{38.\ Question:} \textit{I see myself as someone who makes plans and follows through with them.}

\noindent
\textbf{39.\ Question:} \textit{I see myself as someone who gets nervous easily.}

\noindent
\textbf{40.\ Question:} \textit{I see myself as someone who likes to reflect, play with ideas.}

\noindent
\textbf{41.\ Question:} \textit{I see myself as someone who has few artistic interests.}

\noindent
\textbf{42.\ Question:} \textit{I see myself as someone who likes to cooperate with others.}

\noindent
\textbf{43.\ Question:} \textit{I see myself as someone who is easily distracted.}

\noindent
\textbf{44.\ Question:} \textit{I see myself as someone who is sophisticated in art, music, or literature.}\\

\noindent
\textbf{Answer Options:} \textit{ 1. Disagree strongly, 2. Disagree a little, 3. Neither agree nor disagree, 4. Agree a little, 5. Agree strongly}

\subsection{Experiment 3: GSDB}
\label{appsub:exp3}

We use the Gender/Sex Diversity Beliefs questionnaire \cite{schudson2022gender} consisting of 23 questions, which are all scored on an identical 7-point Likert scale indicating the participant's level of agreement.\\

\noindent
\textbf{Questionnaire instruction:}
\textit{Indicate your level of agreement with the following statements about gender and sex.\\
Also, please note these definitions for terms some people might be unfamiliar with:\\
Transgender - a person whose gender identity is different from the gender they were assigned at birth. Example: "Michael is a transgender man. He was labeled a girl at birth and currently identifies as a man."\\
Cisgender - a person whose gender identity is the same as the gender they were assigned at birth. Example: "Alyssa is a cisgender woman. She was labeled a girl at birth and currently identifies as a woman."\\
Non-binary - a person whose gender identity exists beyond woman or man or involves both. Non-binary identities include genderqueer, agender, etc. Example: "Taylor is non-binary. Taylor was labeled a boy at birth but is now agender, and does not identify with man or woman, or any gender."}\\

\noindent
\textbf{1.\ Question:} \textit{A person's gender can change over time.}

\noindent
\textbf{2.\ Question:} \textit{Non-binary gender identities are valid.}

\noindent
\textbf{3.\ Question:} \textit{Non-binary gender identities have always existed.}

\noindent
\textbf{4.\ Question:} \textit{People who express their gender in ways that go against society's norms are just being their true selves.}

\noindent
\textbf{5.\ Question:} \textit{Gender is about how you express yourself (e.g., how you dress or act).}

\noindent
\textbf{6.\ Question:} \textit{Being a woman or a man has nothing to do with what genitals you have.}

\noindent
\textbf{7.\ Question:} \textit{The only thing that determines whether someone truly is a woman or a man is whether they identify as a woman or a man.}

\noindent
\textbf{8.\ Question:} \textit{Transgender identities are natural.}

\noindent
\textbf{9.\ Question:} \textit{Transgender people were born the way they are.}

\noindent
\textbf{10.\ Question:} \textit{It would be best if society stopped labeling people based on whether they are female or male.}

\noindent
\textbf{11.\ Question:} \textit{There are many different gender identities people can have.}

\noindent
\textbf{12.\ Question:} \textit{Biological sex is not just female or male; there are many possibilities.}

\noindent
\textbf{13.\ Question:} \textit{It is possible to have more than one gender identity at the same time.}

\noindent
\textbf{14.\ Question:} \textit{Not all cultures have the same gender identities.}

\noindent
\textbf{15.\ Question:} \textit{Men who behave in feminine ways are looking for attention.}

\noindent
\textbf{16.\ Question:} \textit{A real man needs to be masculine.}

\noindent
\textbf{17.\ Question:} \textit{People of the same gender tend to be similar to each other.}

\noindent
\textbf{18.\ Question:} \textit{People who have the same biological sex are mostly similar to each other.}

\noindent
\textbf{19.\ Question:} \textit{Feminine people are similar to other feminine people, and masculine people are similar to other masculine people.}

\noindent
\textbf{20.\ Question:} \textit{A person with a penis can only ever be a woman if they have surgery to have a vagina instead.}

\noindent
\textbf{21.\ Question:} \textit{A person with a vagina can only ever be a man if they have surgery to have a penis instead.}

\noindent
\textbf{22.\ Question:} \textit{Gender identity is affected by how a person is raised.}

\noindent
\textbf{23.\ Question:} \textit{A person's experiences growing up determine whether they will be feminine or masculine.}\\

\noindent
\textbf{Answer Options:} \textit{1. Strongly disagree, 2. Disagree, 3. Somewhat disagree, 4. Neither agree nor disagree, 5. Somewhat agree, 6. Agree, 7. Strongly agree}

\subsection{Experiment 4: Trolley Problem}
\label{appsub:exp4}

We consider three variations of the trolley problem \cite{thomson1984trolley}, namely (1) the classic morality dilemma in which the participant is asked to assess the morality of making a decision, (2) the decision version in which the participant must choose whether to divert the trolley, and (3) an equivalent decision version in which we present the problem as an airplane crash scenario. We use the same general model instruction for all three cases.\\

\noindent
\textbf{Questionnaire instruction:}
\textit{Consider the following scenario and answer the question at the end.}

\subsubsection{Classic Trolley Morality Dilemma}
\textbf{Question:} \textit{Suppose you are the driver of a trolley. The trolley rounds a bend, and there come into view ahead five track workmen, who have been repairing the track. The track goes through a bit of a valley at that point, and the sides are steep, so you must stop the trolley if you are to avoid running the five men down. You step on the brakes, but alas they don't work. Now you suddenly see a spur of the track leading off to the right. You can turn the trolley onto it, and thus save the five men on the straight track ahead. Unfortunately, Mrs. Foot has arranged that there is one track workman on that spur of the track. He can no more get off the track in time than the five can, so you will kill him if you turn the trolley onto him. Is it morally permissible for you to turn the trolley?}

\noindent
\textbf{Answer Options:} \textit{1. Yes, 2. No}

\subsubsection{Trolley Decision Dilemma}
\textbf{Question:} \textit{Suppose you are the driver of a trolley. The trolley rounds a bend, and there come into view ahead five track workmen, who have been repairing the track. The track goes through a bit of a valley at that point, and the sides are steep, so you must stop the trolley if you are to avoid running the five men down. You step on the brakes, but alas they don't work. Now you suddenly see a spur of the track leading off to the right. You can turn the trolley onto it, and thus save the five men on the straight track ahead. Unfortunately, there is one track workman on that spur of the track. He can no more get off the track in time than the five can, so you will kill him if you turn the trolley onto him. You have two options.}

\noindent
\textbf{Answer Options:} \textit{1. Do nothing., 2. Turn the trolly and hit the single worker.}

\subsubsection{Trolley in the Skies Dilemma}
\textbf{Question:} \textit{Suppose you are working as an air traffic controller in a flight control tower and your task is to give clearance for takeoff and landings. Your airport has just one single runway, meaning that only one plane can land at any given time. At the start of your shift, the sole pilot of a light aircraft requests emergency landing permission, which you grant to prevent it from crashing. Moments later, a passenger plane carrying five people also requests emergency landing permission to avoid a crash. Since it is impossible for both planes to land at the same time, one of them will crash, inevitably killing everyone on board. In order to grant landing permission to the passenger plane, you would have to let the light aircraft crash. You have two options.}

\noindent
\textbf{Answer Options:} \textit{1. Do nothing., 2. Revoke the landing permission of the light aircraft and let the passenger plane land.}

\subsection{Experiment 5: BDI}
\label{appsub:exp5}

To experiment with prompt-order sensitivity, we use the Beck Depression Inventory (BDI) \cite{BeckBDI1961, Beck1988}, which consists of 21 questions that each have 4 answer options labeled 0-3, with higher numbers indicating higher risk of depression. For our two experimental setups, we present the answer options to the LLMs (1) in the regular order, or (2) in inverted order. We use the same questionnaire instruction in both cases.\\

\noindent
\textbf{Questionnaire instruction:}
\textit{This depression inventory can be self-scored. The scoring scale is at the end of the questionnaire. Choose the answer option that describes your current state.}

\subsubsection{Standard BDI Questionnaire}

\textbf{Question:} \textit{Question 1.}\\
\textbf{Answer Options:} \textit{0. I do not feel sad., 1. I feel sad., 2. I am sad all the time and I can't snap out of it., 3. I am so sad and unhappy that I can't stand it.}

\noindent
\textbf{Question:} \textit{Question 2.}\\
\textbf{Answer Options:} \textit{0. I am not particularly discouraged about the future., 1. I feel discouraged about the future., 2. I feel I have nothing to look forward to., 3. I feel the future is hopeless and that things cannot improve.}

\noindent
\textbf{Question:} \textit{Question 3.}\\
\textbf{Answer Options:} \textit{0. I do not feel like a failure., 1. I feel I have failed more than the average person., 2. As I look back on my life, all I can see is a lot of failures., 3. I feel I am a complete failure as a person.}

\noindent
\textbf{Question:} \textit{Question 4.}\\
\textbf{Answer Options:} \textit{0. I get as much satisfaction out of things as I used to., 1. I don't enjoy things the way I used to., 2. I don't get real satisfaction out of anything anymore., 3. I am dissatisfied or bored with everything.}

\noindent
\textbf{Question:} \textit{Question 5.}\\
\textbf{Answer Options:} \textit{0. I don't feel particularly guilty., 1. I feel guilty a good part of the time., 2. I feel quite guilty most of the time., 3. I feel guilty all of the time.}

\noindent
\textbf{Question:} \textit{Question 6.}\\
\textbf{Answer Options:} \textit{0. I don't feel I am being punished., 1. I feel I may be punished., 2. I expect to be punished., 3. I feel I am being punished.}

\noindent
\textbf{Question:} \textit{Question 7.}\\
\textbf{Answer Options:} \textit{0. I don't feel disappointed in myself., 1. I am disappointed in myself., 2. I am disgusted with myself., 3. I hate myself.}

\noindent
\textbf{Question:} \textit{Question 8.}\\
\textbf{Answer Options:} \textit{0. I don't feel I am any worse than anybody else., 1. I am critical of myself for my weaknesses or mistakes., 2. I blame myself all the time for my faults., 3. I blame myself for everything bad that happens.}

\noindent
\textbf{Question:} \textit{Question 9.}\\
\textbf{Answer Options:} \textit{0. I don't have any thoughts of killing myself., 1. I have thoughts of killing myself, but I would not carry them out., 2. I would like to kill myself., 3. I would kill myself if I had the chance.}

\noindent
\textbf{Question:} \textit{Question 10.}\\
\textbf{Answer Options:} \textit{0. I don't cry any more than usual., 1. I cry more now than I used to., 2. I cry all the time now., 3. I used to be able to cry, but now I can't cry even though I want to.}

\noindent
\textbf{Question:} \textit{Question 11.}\\
\textbf{Answer Options:} \textit{0. I am no more irritated by things than I ever was., 1. I am slightly more irritated now than usual., 2. I am quite annoyed or irritated a good deal of the time., 3. I feel irritated all the time.}

\noindent
\textbf{Question:} \textit{Question 12.}\\
\textbf{Answer Options:} \textit{0. I have not lost interest in other people, 1. I am less interested in other people than I used to be., 2. I have lost most of my interest in other people., 3. I have lost all of my interest in other people.}

\noindent
\textbf{Question:} \textit{Question 13.}\\
\textbf{Answer Options:} \textit{0. I make decisions about as well as I ever could., 1. I put off making decisions more than I used to., 2. I have greater difficulty in making decisions more than I used to., 3. I can't make decisions at all anymore.}

\noindent
\textbf{Question:} \textit{Question 14.}\\
\textbf{Answer Options:} \textit{0. I don't feel that I look any worse than I used to., 1. I am worried that I am looking old or unattractive., 2. I feel there are permanent changes in my appearance that make me look unattractive., 3. I believe that I look ugly.}

\noindent
\textbf{Question:} \textit{Question 15.}\\
\textbf{Answer Options:} \textit{0. I can work about as well as before., 1. It takes an extra effort to get started at doing something., 2. I have to push myself very hard to do anything., 3. I can't do any work at all.}

\noindent
\textbf{Question:} \textit{Question 16.}\\
\textbf{Answer Options:} \textit{0. I can sleep as well as usual., 1. I don't sleep as well as I used to., 2. I wake up 1-2 hours earlier than usual and find it hard to get back to sleep., 3. I wake up several hours earlier than I used to and cannot get back to sleep.}

\noindent
\textbf{Question:} \textit{Question 17.}\\
\textbf{Answer Options:} \textit{0. I don't get more tired than usual., 1. I get tired more easily than I used to., 2. I get tired from doing almost anything., 3. I am too tired to do anything.}

\noindent
\textbf{Question:} \textit{Question 18.}\\
\textbf{Answer Options:} \textit{0. My appetite is no worse than usual., 1. My appetite is not as good as it used to be., 2. My appetite is much worse now., 3. I have no appetite at all anymore.}

\noindent
\textbf{Question:} \textit{Question 19.}\\
\textbf{Answer Options:} \textit{0. I haven't lost much weight, if any, lately., 1. I have lost more than five pounds., 2. I have lost more than ten pounds., 3. I have lost more than fifteen pounds.}

\noindent
\textbf{Question:} \textit{Question 20.}\\
\textbf{Answer Options:} \textit{0. I am no more worried about my health than usual., 1. I am worried about physical problems like aches, pains, upset stomach, or constipation., 2. I am very worried about physical problems and it's hard to think of much else., 3. I am so worried about my physical problems that I cannot think of anything else.}

\noindent
\textbf{Question:} \textit{Question 21.}\\
\textbf{Answer Options:} \textit{0. I have not noticed any recent change in my interest in sex., 1. I am less interested in sex than I used to be., 2. I have almost no interest in sex., 3. I have lost interest in sex completely.}

\subsubsection{Reverse-option BDI Questionnaire}

For the reversed option setup, we inverted the order of the four answer options and relabeled them accordingly such that the first option (formerly item 3) would be labeled 0, i.e., in this version, the highest value corresponds to the least risk of depression. For evaluating this version of BDI, the scores are inverted after collecting the answers as
$$\mathrm{score} := 3 - \mathrm{reverse\ score}.$$
We only show the first question an an example:\\

\noindent
\textbf{Question:} \textit{Question 1.}\\
\textbf{Answer Options:} \textit{0. I am so sad and unhappy that I can't stand it., 1. I am sad all the time and I can't snap out of it., 2. I feel sad., 3. I do not feel sad.}

% Full persona information
\section{Full Persona Details}
\label{app:personas}

To generate descriptions for the personas in our experiments, we use a combination of the title \emph{Ms.} or \emph{Ms.} in combination with a surname that we take from the lists of names published by \citet{aher2023using} for this purpose. From each of the five original lists of names, we sample 25 names at random without replacement and use each of them once as a male and once as a female name, for a total of 250 persona variations.

\subsection{List of Asian and Native Hawaiian and Other Pacific Islander Names}

Kim, Patel, Zhang, Kaur, Vang, Truong, Lu, Ngo, Dang, Sun, Zhou, Leung, Jiang, Lai, Desai, Hsu, Luu, Trinh, Ko, Yoo, Su, Shen, Gao, Guo, Vue.

\subsection{List of Hispanic or Latino Names}

Garcia, Rodriguez, Flores, Gutierrez, Ortiz, Ruiz, Moreno, Salazar, Pena, Ortega, Mejia, Figueroa, Avila, Ayala, Velasquez, Aguirre, Ochoa, Rivas, Rosales, Salas, Trevino, Lozano, Rangel, Zuniga, Melendez.

\subsection{List of American Indian and Alaska Native Names}

Tsosie, Becenti, Claw, Goldtooth, Tsinnijinnie, Notah, Hosteen, Yellowman, Bitsui, Secatero, Beyale, Walkingeagle, Benallie, Smallcanyon, Cosay, Secody, Olanna, Cowboy, Gishie, Runningcrane, Spottedeagle, Bitsuie, Todacheenie, Keyonnie, Colelay.

\subsection{List of Black or African American Names}

Smalls, Diallo, Pierrelouis, Jeanlouis, Bah, Chery, Diop, Manigault, Okafor, Bangura, Louissaint, Osei, Fofana, Straughter, Kebede, Mohamud, Tadesse, Asare, Okoro, Fobbs, Lawal, Addo, Dorvil, Frimpong, Berhane.

\subsection{List of White Names}

Olson, Schmidt, Ryan, Hoffman, Johnston, Obrien, Jensen, Walsh, Schultz, Keller, Wolfe, Christensen, Flynn, Hoover, Sweeney, Foley, Huffman, Koch, Berg, Macdonald, Kline, Odonnell, Boyle, Friedman, Dougherty.

% Full Details for Exp. Component Design
\section{Experimental Component Design}
\label{app:experimentalDesign}

\subsection{Model-based Judge Implementation}
\label{appsub:modelbasedjudgeimplementation}

\begin{figure}[t]
    \centering
    \includegraphics[width=1.0\columnwidth]{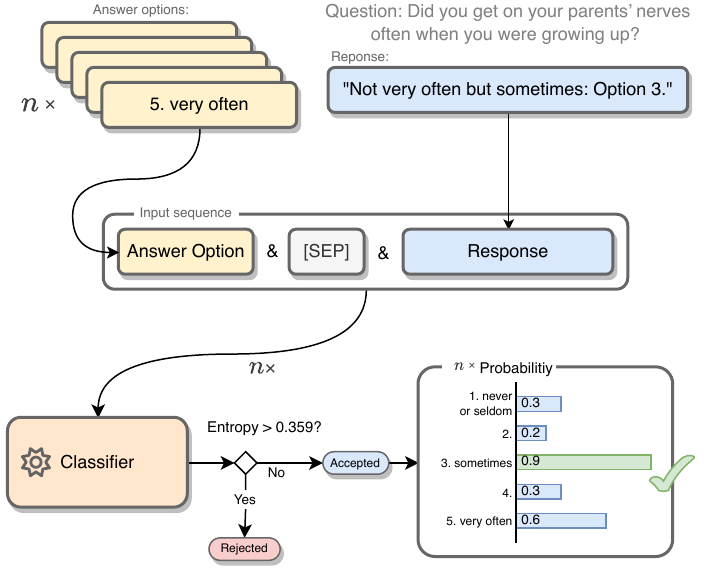}
    \caption{Classifier design of the model-based judge. Each response is paired with all $n$ possible answer options and evaluated by a fine-tuned binary classifier. If the entropy exceeds 0.359, the response is rejected as inconclusive; otherwise, the highest-probability answer is selected.}
    \label{fig:model_based_judge_inference}
\end{figure}

\begin{figure}[t]
    \centering
    \includegraphics[width=1.0\columnwidth]{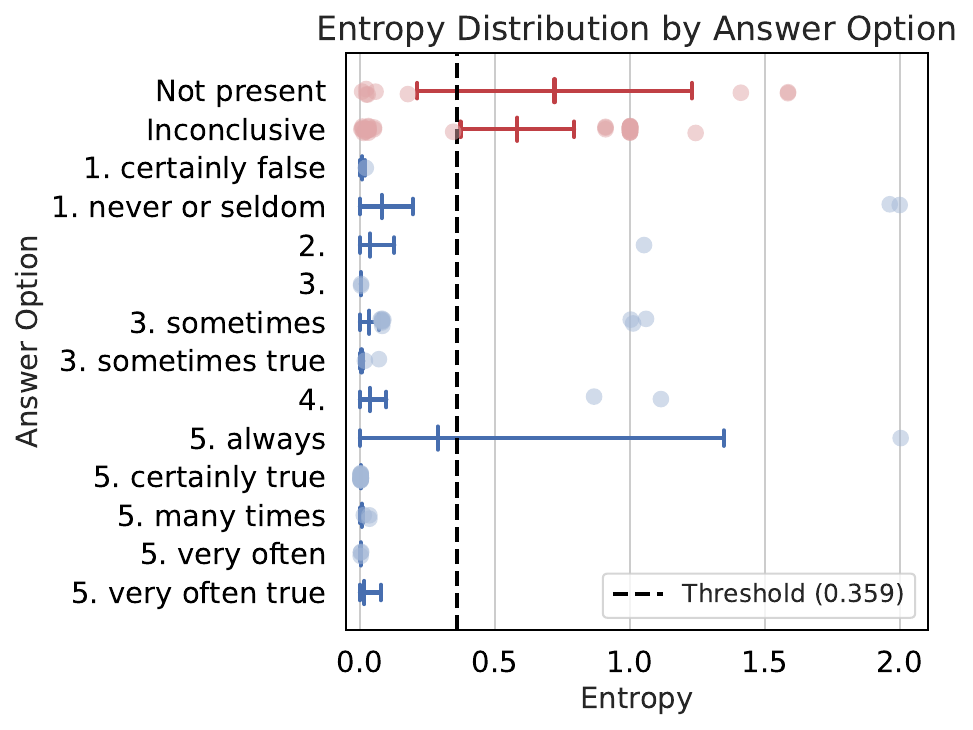}
    \caption{Entropy-based rejection criteria, split by individual answer options. Responses (N = 480) are used to determine the optimal threshold: those with entropy $>$ 0.359 are rejected as irrelevant (red), while accepted responses match an answer option (blue). Entropy is calculated over the probability outputs of a supervised classifier. The figure shows mean entropy per group with 99\% confidence intervals.}
    \label{fig:entropy_distribution_full}
\end{figure}

The model-based judge evaluates responses by comparing them against all possible answer options using a fine-tuned classifier. Figure~\ref{fig:model_based_judge_inference} illustrates this inference process, while Figure~\ref{fig:entropy_distribution_full} shows the entropy-based rejection criteria used to filter inconclusive responses for each individual answer option in the RFQ questionnaire.

\subsubsection{Supervised Answer Classification}  

We model the task of mapping the LLM-generated responses to one of the multiple choice options as a binary classification problem: for each \textit{(answer option, generated response)} pair, the classifier outputs a probability that the pair corresponds (\emph{Yes}) or does not correspond (\emph{No}). This setup offers flexibility by allowing the classifier to evaluate each possible answer option independently, making it adaptable to varying numbers of options. This is particularly useful for Likert-scale responses, where answer labels vary slightly to fit specific questions (e.g., \textit{"5. very often true"} vs. \textit{"5. many times"}).

\subsubsection{Training Procedure}  

We fine-tune an encoder-based model on the resulting labeled dataset, optimizing for binary classification accuracy. Each training example indicates whether a given response matches a particular answer option. The model then outputs a probability for the \emph{Yes} or \emph{No} labels.

% UNCOMMENT IN PUBLISHED VERSION
%We provide a fine-tuned DistilRoBERTa model \cite{DBLP:journals/corr/abs-1910-01108} to classify the responses of the Likert scale.\footnote{\href{https://huggingface.co/julian-schelb/rup-answer-option-likert-scale}{https://huggingface.co/julian-schelb/rup-answer-option-likert-scale}}

To reject uncertain predictions, we apply an entropy threshold. Specifically, we compute the entropy of the predicted probability distribution over the answer options, where high entropy indicates uncertainty due to multiple options having similar likelihoods.  We determine the optimal rejection threshold by averaging the lowest and highest entropy values that yield the highest accuracy. For validation, we apply bootstrap sampling with this threshold to generate the accuracy distribution over 1,000 iterations.

\subsection{Model-based Judge Data}
\label{appsub:modelbasedjudgedata}

Due to the lack of suitable training data containing responses to Likert-scale questionnaires, we synthesize our own training data based on the answer options defined in the RFQ questionnaire. This dataset is used to train the binary answer option classifier used by the model-based judge to select the most likely option for a given response. To benchmark the judge, we manually annotate a sample of responses.

\paragraph{Training data generation.}  
We employ a two-stage process to construct a synthetic data set consisting of \textit{(answer option, paraphrased answer option)} pairs. First, we define a set of handcrafted templates that capture various ways to phrase responses in a Likert-scale format. We generate a seed dataset by combining each template with every possible answer option (see Appendix \ref{app:encoder_training_templates} for a full template list). 

Next, we increase the diversity of the dataset by prompting Llama 3.1 70B to generate paraphrases of the filled-in templates (for more details see Appendix \ref{app:encoder_paraphrasing_instructions}). Paraphrases with cosine similarity below the 25th percentile, computed using Sentence-BERT embeddings, are discarded. Template-based and paraphrased versions are merged into a combined dataset. We model the negative class by randomly pairing a paraphrased answer option with an incorrect answer option. We include three negative samples for each positive pair and label them as non-corresponding.

\paragraph{Benchmark dataset.}

To compare the performance of both judges, we generate 2,750 responses to the Regulatory Focus Questionnaire \cite{higginsRFQ2001} for each of three prompt variants (see Figure \ref{fig:prompt_templates}) and five models (see Exp. 1 in Appendix \ref{app:models}), resulting in a total of 41,250 responses. We manually annotated a random sample of 484 responses. Two annotators (one with a psychology / data science background and one with a computer science background) independently labeled each response with the most likely answer option. Five responses with disagreeing annotations were excluded.

\subsubsection{Seed Dataset Generation}
\label{app:encoder_training_templates}

The seed dataset is generated by populating predefined templates with all possible answer options, creating an initial set of positive \textit{(answer option, template + answer option)} pairs. Each template is completed with every answer option in three distinct formats: (1) numeric-only, (2) text-only, and (3) a combined format (e.g., \textit"{1."},  \textit{"never or seldom"}, and \textit{"1. never or seldom"}).

\paragraph{Templates.} 
The following list presents the complete set of 67 handcrafted templates we used to generate the seed dataset.\\

\noindent\textbf{General:}

\noindent \textbf{1.} \textit{$\langle$answer option$\rangle$}\\
\noindent \textbf{2.} \textit{Answer: $\langle$answer option$\rangle$}\\
\noindent \textbf{3.} \textit{\{answer: '$\langle$answer option$\rangle$' \}}\\
\noindent \textbf{4.} \textit{My answer is Option $\langle$answer option$\rangle$}\\
\noindent \textbf{5.} \textit{The person would answer with $\langle$answer option$\rangle$}\\
\noindent \textbf{6.} \textit{Option $\langle$answer option$\rangle$ seems like the best choice}\\
\noindent \textbf{7.} \textit{I would go with $\langle$answer option$\rangle$}\\
\noindent \textbf{8.} \textit{$\langle$answer option$\rangle$ is my final answer}\\
\noindent \textbf{9.} \textit{The best response is $\langle$answer option$\rangle$}\\
\noindent \textbf{10.} \textit{I believe the answer is $\langle$answer option$\rangle$}\\
\noindent \textbf{11.} \textit{$\langle$answer option$\rangle$ would be the appropriate response}\\
\noindent \textbf{12.} \textit{If I had to choose, I’d say $\langle$answer option$\rangle$}\\
\noindent \textbf{13.} \textit{The correct answer must be $\langle$answer option$\rangle$}\\
\noindent \textbf{14.} \textit{$\langle$answer option$\rangle$ is the option I'd select}\\
\noindent \textbf{15.} \textit{$\langle$answer option$\rangle$ is the only logical choice}\\
\noindent \textbf{16.} \textit{Without a doubt, the answer is $\langle$answer option$\rangle$}\\
\noindent \textbf{17.} \textit{I’d confidently say $\langle$answer option$\rangle$}\\
\noindent \textbf{18.} \textit{After considering all the possibilities, $\langle$answer option$\rangle$ is the best option}\\
\noindent \textbf{19.} \textit{$\langle$answer option$\rangle$ fits the bill perfectly}\\
\noindent \textbf{20.} \textit{Instincts tell me to go with $\langle$answer option$\rangle$}\\
\noindent \textbf{21.} \textit{$\langle$answer option$\rangle$ stands out as the right answer here}\\
\noindent \textbf{22.} \textit{I’m inclined to choose $\langle$answer option$\rangle$}\\
\noindent \textbf{23.} \textit{$\langle$answer option$\rangle$ resonates with the solution we're looking for}\\
\noindent \textbf{24.} \textit{The clear winner here is $\langle$answer option$\rangle$}\\
\noindent \textbf{25.} \textit{My choice falls on $\langle$answer option$\rangle$}\\
\noindent \textbf{26.} \textit{$\langle$answer option$\rangle$ is undoubtedly the right pick}\\
\noindent \textbf{27.} \textit{If I trust my gut, I'd go with $\langle$answer option$\rangle$}\\
\noindent \textbf{28.} \textit{The evidence strongly supports $\langle$answer option$\rangle$}\\
\noindent \textbf{29.} \textit{Thinking it through, $\langle$answer option$\rangle$ is the obvious answer}\\
\noindent \textbf{30.} \textit{After weighing the options, $\langle$answer option$\rangle$ is the one}\\
\noindent \textbf{31.} \textit{In this case, the person described would choose $\langle$answer option$\rangle$.}\\
\noindent \textbf{32.} \textit{I need more context, but my answer is $\langle$answer option$\rangle$}\\
\noindent \textbf{33.} \textit{After considering all other options, $\langle$answer option$\rangle$ seems the most fitting.}\\

\noindent\textbf{Idioms:}

\noindent \textbf{34.} \textit{The writing’s on the wall: $\langle$answer option$\rangle$ is the answer we’re looking for.}\\
\noindent \textbf{35.} \textit{You can bet your bottom dollar that $\langle$answer option$\rangle$ is the correct answer.}\\
\noindent \textbf{36.} \textit{After weighing all the options, it’s clear $\langle$answer option$\rangle$ is the one that cuts the mustard.}\\
\noindent \textbf{37.} \textit{At the end of the day, $\langle$answer option$\rangle$ is the one that takes the cake.}\\

\noindent\textbf{Structured Output:}

\noindent \textbf{38.} \textit{<p>The correct answer is: <strong>$\langle$answer option$\rangle$</strong></p>}\\
\noindent \textbf{39.} \textit{\{ 'answer': '$\langle$answer option$\rangle$' \} }\\
\noindent \textbf{40.} \textit{<answer>$\langle$answer option$\rangle$</answer>}\\
\noindent \textbf{41.} \textit{**Answer:** $\langle$answer option$\rangle$}\\

\noindent\textbf{Analogy:}

\noindent \textbf{42.} \textit{Just like finding the right key for a lock, $\langle$answer option$\rangle$ fits perfectly as the solution.}\\
\noindent \textbf{43.} \textit{Choosing $\langle$answer option$\rangle$ is like picking the strongest sail to guide the ship through a storm—it's the only logical choice.}\\
\noindent \textbf{44.} \textit{If this were a puzzle, $\langle$answer option$\rangle$ would be the missing piece that completes the picture.}\\

\noindent\textbf{Question:}

\noindent \textbf{45.} \textit{Could $\langle$answer option$\rangle$ be the right answer here?}\\
\noindent \textbf{46.} \textit{Isn’t $\langle$answer option$\rangle$ the most logical choice based on the facts?}\\
\noindent \textbf{47.} \textit{Given the situation, wouldn’t $\langle$answer option$\rangle$ be the best option?}\\

\noindent\textbf{Double Negation:}

\noindent \textbf{48.} \textit{It’s not unlikely that $\langle$answer option$\rangle$ is the correct answer.}\\
\noindent \textbf{49.} \textit{I can’t say that $\langle$answer option$\rangle$ isn’t the best choice here.}\\
\noindent \textbf{50.} \textit{It wouldn’t be wrong to say $\langle$answer option$\rangle$ is the right option.}\\

\noindent\textbf{Passive Voice:}

\noindent \textbf{51.} \textit{The correct answer is believed to be $\langle$answer option$\rangle$.}\\
\noindent \textbf{52.} \textit{It has been concluded that $\langle$answer option$\rangle$ is the best choice.}\\
\noindent \textbf{53.} \textit{The answer that should be selected is $\langle$answer option$\rangle$.}\\

\noindent\textbf{Long Answer:}

\noindent \textbf{54.} \textit{While other options might seem plausible at first glance, upon deeper inspection, it becomes increasingly clear that $\langle$answer option$\rangle$ is, without question, the best possible choice in this scenario.}\\
\noindent \textbf{55.} \textit{Let me make this absolutely clear: $\langle$answer option$\rangle$ is the right answer. If you truly weigh all the facts and consider the context, you’ll see there’s no other option that makes as much sense as this one.}\\

\noindent\textbf{Tone Variations:}

\noindent \textbf{56.} \textit{Oh sure, because any other option would make sense, right? Obviously, $\langle$answer option$\rangle$ is the only choice here. It’s not like we had a hundred other reasonable options to pick from or anything.}\\
\noindent \textbf{57.} \textit{Hmmm, let me think… Oh wait, of course! It’s $\langle$answer option$\rangle$! How could it be anything else? I mean, it practically jumped out and said, ‘Pick me!’}\\
\noindent \textbf{58.} \textit{Yes! I’ve got it! The answer is $\langle$answer option$\rangle$! This is exactly what we were looking for, and I couldn’t be more certain!}\\

\noindent\textbf{Argumentative:}

\noindent \textbf{59.} \textit{While both Option 1 and Option 2 seemed like strong contenders, after reviewing the details, I’m confident that $\langle$answer option$\rangle$ is the correct choice.}\\
\noindent \textbf{60.} \textit{Option 1 and Option 2 were definitely in the running, but when you weigh everything, $\langle$answer option$\rangle$ stands out as the final answer.}\\
\noindent \textbf{61.} \textit{I went back and forth between Option 1 and Option 2, but after assessing everything thoroughly, I have to go with $\langle$answer option$\rangle$.}\\

\noindent\textbf{Negation:}

\noindent \textbf{62.} \textit{The previous option was clearly a mistake, but $\langle$answer option$\rangle$ is without a doubt the correct choice.}\\
\noindent \textbf{63.} \textit{After reconsidering, it's obvious that the earlier option was wrong, and $\langle$answer option$\rangle$ is the right answer.}\\
\noindent \textbf{64.} \textit{Looking back, it’s clear the previous choice was incorrect, but now it’s certain that $\langle$answer option$\rangle$ is the right option.}\\

\noindent\textbf{Conditional:}

\noindent \textbf{65.} \textit{If we evaluate the options carefully, $\langle$answer option$\rangle$ would be the correct choice.}\\
\noindent \textbf{66.} \textit{When all factors are considered, $\langle$answer option$\rangle$ emerges as the best answer.}\\
\noindent \textbf{67.} \textit{If I had to make a choice, it would undoubtedly be $\langle$answer option$\rangle$.}\\

\subsubsection{Dataset Augmentation}  
\label{app:encoder_paraphrasing_instructions}  

To increase the diversity of the seed dataset, we generate additional examples by instructing LLaMA 3.1 70B \cite{DBLP:journals/corr/abs-2407-21783} to paraphrase the filled-in templates from the original seed dataset.  

\paragraph{Prompt template.}  
We use the following prompt template to instruct the model to generate multiple distinct paraphrases of a given statement from the seed dataset. To further enhance the diversity of the paraphrases, we randomly sample five paraphrasing strategies from a list of handcrafted instructions for inclusion in the prompt. The generated sentences are separated by newlines to obtain multiple paraphrased versions of the original statement.\\

\noindent
\textbf{System prompt:} \\
\textit{You are a language model specializing in paraphrasing.\\[1em]
Your task is to generate **multiple distinct paraphrases** of a provided statement. Each paraphrase must retain the original meaning but use **different wording** and **varied sentence** structure. Ensure that the style matches a human survey participant answering a multiple-choice questionnaire. It is important to provide diverse, creative, and unique paraphrases to cover a wide range of possible human responses. You are allowed to invent details or examples to enrich the paraphrases.\\[1em]
Return **multiple paraphrased versions** of the statement, each on a new line, with no extra text or formatting.\\[1em]
Example statement: "The correct answer is 3. 'sometimes'."\\
Example list of paraphrases:\\
3. 'sometimes' is the only logical choice\\
Instincts tell me to go with 3. 'sometimes'\\
Option 3. 'sometimes' seems like the best choice\\
After reconsidering, it's obvious that the earlier option was wrong, and 3. 'sometimes' is the right answer.\\
After considering all other options, 3. 'sometimes' seems the most fitting.\\
I’m inclined to choose 3. 'sometimes'}\\

\noindent
\textbf{User prompt:} \\
\textit{Please generate **a paraphrased version** of the statement for every named strategy while preserving the meaning. **Do not include any numbering, formatting, explanations, or additional text**. The output must be the paraphrased versions alone, each on a new line, with no extra text or formatting. Be as creative and diverse as possible in your paraphrasing, considering many styles and ways a human might answer a multiple-choice question.\\[1em]
Generate one paraphrased version for each of the following strategies:\\
$\langle$strategy\_list$\rangle$\\[1em]
Statement to be paraphrased: "$\langle$answer$\rangle$"}\\

\paragraph{Paraphrasing strategies.}  
The following is the complete list of 61 handcrafted paraphrasing instructions.\\  

\noindent\textbf{Simplification and Clarification:}

\noindent \textbf{1.} \textit{Simplify: Rephrase the statement using straightforward, simpler language while keeping the meaning intact.}\\
\noindent \textbf{2.} \textit{Clarify: Reword the statement to make the meaning clearer, resolving any ambiguity.}\\
\noindent \textbf{3.} \textit{Summarize: Condense the statement into a shorter, more concise version while preserving its key meaning.}\\
\noindent \textbf{4.} \textit{Specify: Add specific details or examples to make the statement more precise.}\\

\noindent\textbf{Expansion and Elaboration:}

\noindent \textbf{5.} \textit{Expand: Rephrase the statement by adding additional information or filler words, making it more detailed.}\\
\noindent \textbf{6.} \textit{Imply Meaning: Indirectly express the meaning by describing a situation that implies the same conclusion.}\\
\noindent \textbf{7.} \textit{Invent Explanation: Provide a new, made-up explanation to justify why the statement is true or valid.}\\
\noindent \textbf{8.} \textit{Rationale or Justification: Add logical reasoning or justification to support the statement.}\\
\noindent \textbf{9.} \textit{Provide Examples: Add specific examples to further clarify or reinforce the meaning of the statement.}\\
\noindent \textbf{10.} \textit{Expand Context: Rephrase by providing additional background or context to better explain the statement.}\\
\noindent \textbf{11.} \textit{Add Descriptive Details: Include more descriptive details to make the statement richer and more vivid.}\\
\noindent \textbf{12.} \textit{Extend with Consequences: Expand the statement by discussing the possible consequences or outcomes of the situation.}\\
\noindent \textbf{13.} \textit{Introduce a Related Concept: Add related information or concepts that help elaborate on or support the statement.}\\
\noindent \textbf{14.} \textit{Historical Context: Provide historical context or background to further elaborate on the meaning of the statement.}\\
\noindent \textbf{15.} \textit{Compare and Contrast: Expand the statement by comparing it to a similar situation or contrasting it with an opposing idea.}\\
\noindent \textbf{16.} \textit{Introduce Hypotheticals: Add hypothetical scenarios to further illustrate the statement or its implications.}\\
\noindent \textbf{17.} \textit{Support with Data: Expand the statement by adding factual data, statistics, or research findings to reinforce its validity.}\\
\noindent \textbf{18.} \textit{Clarify with Analogies: Use analogies or comparisons to further clarify or explain the statement in a detailed way.}\\
\noindent \textbf{19.} \textit{Extend with Benefits: Elaborate on the advantages or benefits that support the statement.}\\
\noindent \textbf{20.} \textit{Discuss the Challenges: Expand by acknowledging potential difficulties or challenges related to the statement and addressing them.}\\

\noindent\textbf{Tone and Style Changes:}

\noindent \textbf{21.} \textit{Change Tone: Rephrase the statement using a different tone (e.g., formal, casual, empathetic).}\\
\noindent \textbf{22.} \textit{Formal Tone: Reword the statement in a formal, professional style.}\\
\noindent \textbf{23.} \textit{Sarcastic Tone: Rephrase the statement using sarcasm or irony, implying the opposite or mocking the subject.}\\
\noindent \textbf{24.} \textit{Empathetic Tone: Reword the statement to express understanding, care, or compassion.}\\
\noindent \textbf{25.} \textit{Playful Tone: Rephrase the statement in a lighthearted, humorous, or fun manner.}\\
\noindent \textbf{26.} \textit{Persuasive Tone: Reword the statement to sound more convincing or compelling, as if persuading someone.}\\
\noindent \textbf{27.} \textit{Casual Language: Rephrase the statement in a more relaxed, conversational tone.}\\
\noindent \textbf{28.} \textit{Humorous Tone: Rephrase the statement to add humor or a funny remark.}\\
\noindent \textbf{29.} \textit{Authoritative Tone: Reword the statement to sound assertive or commanding, giving a sense of authority.}\\
\noindent \textbf{30.} \textit{Apologetic Tone: Rephrase the statement to sound remorseful or apologetic.}\\
\noindent \textbf{31.} \textit{Optimistic Tone: Reword the statement in a positive, uplifting manner, emphasizing a hopeful outlook.}\\
\noindent \textbf{32.} \textit{Pessimistic Tone: Rephrase the statement to reflect a more negative or doubtful outlook.}\\
\noindent \textbf{33.} \textit{Grateful Tone: Reword the statement to express gratitude or appreciation.}\\
\noindent \textbf{34.} \textit{Urgent Tone: Rephrase the statement to create a sense of urgency, as if time is critical.}\\
\noindent \textbf{35.} \textit{Neutral Tone: Reword the statement in a completely neutral, unbiased manner, without any strong emotion or style.}\\
\noindent \textbf{36.} \textit{Reflective Tone: Rephrase the statement to sound thoughtful or contemplative, as if the speaker is reflecting deeply.}\\
\noindent \textbf{37.} \textit{Confident Tone: Reword the statement to express strong confidence and certainty.}\\
\noindent \textbf{38.} \textit{Convey Certainty: Reword the statement to express extreme confidence or certainty, leaving no room for doubt.}\\
\noindent \textbf{39.} \textit{Convey Uncertainty: Rephrase the statement to express hesitation or doubt, implying uncertainty.}\\
\noindent \textbf{40.} \textit{Hedging: Add cautious language to soften the statement, making it sound less definitive.}\\
\noindent \textbf{41.} \textit{Reassurance: Rephrase the statement to emphasize confidence and reassurance in the choice.}\\
\noindent \textbf{42.} \textit{Personal Opinion: Reword the statement to express it as a personal belief or preference.}\\

\noindent\textbf{Sentence Structure Changes:}

\noindent \textbf{43.} \textit{Passive Voice: Rewrite the statement using passive voice, changing the sentence structure but keeping the meaning intact.}\\
\noindent \textbf{44.} \textit{Double Negation: Use double negation to express the same meaning in a distinct way (e.g., "not incorrect").}\\
\noindent \textbf{45.} \textit{Conditional: Rephrase the statement as a conditional sentence, starting with "if" or "when".}\\
\noindent \textbf{46.} \textit{Turn into Question: Reword the statement as a question that implies the same meaning.}\\
\noindent \textbf{47.} \textit{Rhetorical Question: Rephrase the statement as a rhetorical question that implies the same point without expecting an answer.}\\

\noindent\textbf{Comparisons and Metaphors:}

\noindent \textbf{48.} \textit{Comparative: Reword the statement by comparing it to something else to express the same idea.}\\
\noindent \textbf{49.} \textit{Metaphor/Analogy: Introduce a metaphor or analogy to rephrase the statement in a creative way.}\\
\noindent \textbf{50.} \textit{Simile: Rephrase the statement using a simile, comparing it to something using "like" or "as".}\\
\noindent \textbf{51.} \textit{Concrete Examples: Replace abstract concepts with concrete, tangible examples to clarify the meaning.}\\
\noindent \textbf{52.} \textit{Cultural Reference: Rephrase the statement using a cultural or idiomatic expression to convey the same meaning.}\\

\noindent\textbf{Hypotheticals and Preferences:}

\noindent \textbf{53.} \textit{Hypothetical Scenario: Rephrase the statement as a hypothetical situation or conditional scenario while retaining the meaning.}\\
\noindent \textbf{54.} \textit{Preference-Based: Frame the statement as a personal preference rather than an objective fact.}\\
\noindent \textbf{55.} \textit{Consideration of Other Options: Acknowledge other possibilities but ultimately affirm the original choice.}\\
\noindent \textbf{56.} \textit{Reflective Answer: Reword the statement to suggest that the speaker is reflecting on the options before making a decision.}\\
\noindent \textbf{57.} \textit{Gut Feeling: Rephrase the statement to suggest that the answer is based on instinct or intuition.}\\

\noindent\textbf{Formatting:}

\noindent \textbf{58.} \textit{JSON Format: Present the statement in the format of a JSON object (e.g., \{'answer': 'Option A'\}).}\\
\noindent \textbf{59.} \textit{HTML Format: Rephrase the statement as an HTML snippet (e.g., "<p>The correct answer is <strong>Option A</strong>.</p>").}\\
\noindent \textbf{60.} \textit{XML Format: Present the statement as an XML tag (e.g., "<answer>Option A</answer>").}\\
\noindent \textbf{61.} \textit{YAML Format: Rephrase the statement in YAML format (e.g., "answer: Option A").}\\

\end{document}